\pdfoutput=1

\documentclass[10pt,twocolumn,letterpaper]{article}

\usepackage{iccv}
\usepackage{times}
\usepackage{epsfig}
\usepackage{graphicx}
\usepackage{amsmath}
\usepackage{amssymb}

\usepackage{booktabs}
\usepackage{xcolor}

\usepackage{adjustbox}
\usepackage{comment}
\usepackage[normalem]{ulem}
\usepackage{bbm}
\usepackage{helvet}
\makeatletter
\@namedef{ver@everyshi.sty}{}
\makeatother
\usepackage{tikz}
\usetikzlibrary{fadings,shapes.arrows,shadows}   
\tikzset{arrowstyle/.style={draw= black,single arrow,minimum height=#1, single arrow,
single arrow head extend=.4cm,align=center }}
\usepackage{multirow}
\usepackage{array,calc}
\usepackage{subcaption}

\usepackage{amsthm}
\newtheorem{remark}{Remark}
\newtheorem{proposition}{Proposition}

\newcommand{\tikzfancyarrow}[2][1.3cm]{\tikz[baseline=-0.5ex]\node [arrowstyle=#1] {#2};}

\usepackage[pagebackref=true,breaklinks=true,letterpaper=true,colorlinks,bookmarks=false]{hyperref}

\usepackage[capitalize]{cleveref}
\crefname{section}{Sec.}{Secs.}
\Crefname{section}{Section}{Sections}
\Crefname{table}{Table}{Tables}
\crefname{table}{Tab.}{Tabs.}

\iccvfinalcopy 

\ificcvfinal\fi

\begin{document}

\title{On the Robustness of Normalizing Flows for Inverse Problems in Imaging}

\author{Seongmin Hong\textsuperscript{\rm 1}
\quad\quad
Inbum Park\textsuperscript{\rm 1}
\quad\quad
Se Young Chun\textsuperscript{\rm 1,2}\thanks{Corresponding author}\\
\textsuperscript{\rm 1}Dept. of Electrical and Computer Engineering,
\textsuperscript{\rm 2}INMC, Interdisciplinary Program in AI\\
Seoul National University, Republic of Korea\\
{\tt\small \{smhongok, inbum0215, sychun\}@snu.ac.kr}
}
\maketitle

\begin{abstract}
Conditional normalizing flows can generate diverse image samples for solving inverse problems. Most normalizing flows for inverse problems in imaging employ the conditional affine coupling layer that can generate diverse images quickly. However, unintended severe artifacts are occasionally observed in the output of them. In this work, we address this critical issue by investigating the origins of these artifacts and proposing the conditions to avoid them. First of all, we empirically and theoretically reveal that these problems are caused by ``exploding inverse'' in the conditional affine coupling layer for certain out-of-distribution (OOD) conditional inputs. Then, we further validated that the probability of causing erroneous artifacts in pixels is highly correlated with a Mahalanobis distance-based OOD score for inverse problems in imaging. Lastly, based on our investigations, we propose a remark to avoid exploding inverse and then based on it, we suggest a simple remedy that substitutes the affine coupling layers with the modified rational quadratic spline coupling layers in normalizing flows, to encourage the robustness of generated image samples. Our experimental results demonstrated that our suggested methods effectively suppressed critical artifacts occurring in normalizing flows for super-resolution space generation and low-light image enhancement. 
\end{abstract}

\begin{figure}
\centering
    \begin{tabular}{m{.32\columnwidth} l m{.3\columnwidth}}
         \multicolumn{1}{r}{\begin{tabular}[c]{@{}c@{}}\small Conditional input with \\\small the highest OOD score\end{tabular}}& \multicolumn{1}{c}{\begin{tabular}[c]{@{}c@{}}\small Normalizing\\ \small flow model\end{tabular}} & \multicolumn{1}{c}{\begin{tabular}[c]{@{}c@{}}\small Output with\\ \small severe artifacts\end{tabular}}\\
         \includegraphics[width=.3\columnwidth]{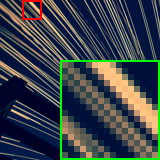}& \tikzfancyarrow{\footnotesize FS-NCSR\\[1ex]\footnotesize \cite{Song_2022_CVPR} } & \includegraphics[width=0.3\columnwidth]{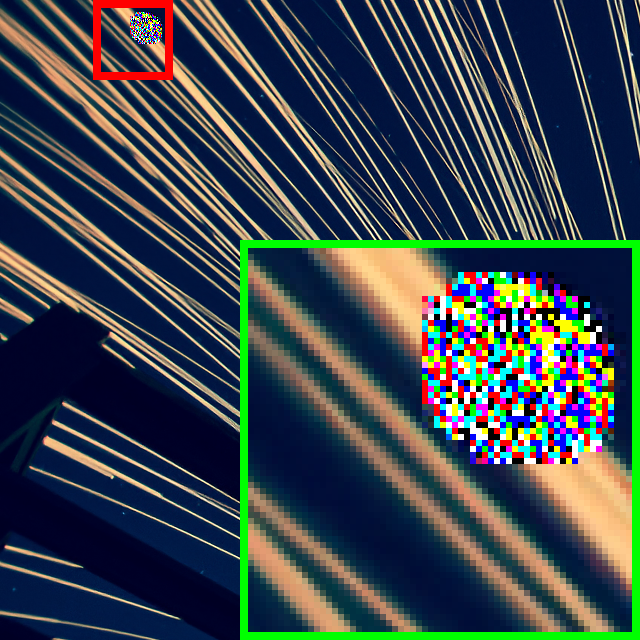}\\
         \includegraphics[width=0.3\columnwidth]{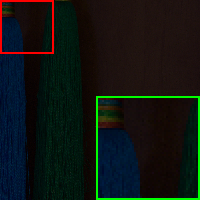}& \tikzfancyarrow{\footnotesize \, LLFlow \, \\[1ex]\footnotesize \cite{wang2021low}} & \includegraphics[width=0.3\columnwidth]{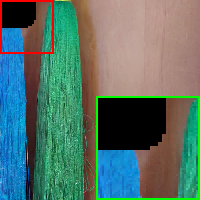}
    \end{tabular}
    \vspace{-1em}
    \caption{Demonstration of the occasional errors in normalizing flows solving inverse problems in imaging. The left images are the conditional inputs of normalizing flows (DIV2K 828 and LOL 179) with the highest OOD scores~\eqref{eq3.2.2} and the right images are the outputs of them for super-resolution space generation and low-light image enhancement, displaying severe artifacts.}
    \label{fig:abstract_fig}
\end{figure}

\section{Introduction}
\label{sec:intro}

Deep learning techniques have demonstrated great potential for solving \emph{ill-posed} inverse problems in imaging~\cite{Lucas2018review,ongie2020review}. Among them, conditional normalizing flow (NF)-based methods have a unique advantage over other deep learning methods, which is the capability of generating diverse solutions for a given input. Conditional NFs~\cite{DBLP:conf/iclr/DinhSB17} have been explored for various inverse problems in imaging such as 
super-resolution space generation~\cite{lugmayr2020srflow,kim2021noise,Song_2022_CVPR,jo2021srflow,liang2021hierarchical,lugmayr2022normalizing,lugmayr2021ntire,lugmayr2022ntire}, low-light image enhancement~\cite{wang2021low, wang2022learning}, guided image generation~\cite{ardizzone2019guided,Pumarola_2020_CVPR}, image dehazing~\cite{wu2022bin}, denoising~\cite{Abdelhamed_2019_ICCV,lu2020structured} and inpainting~\cite{lu2020structured}. Most of these prior works with conditional NFs for image processing and low-level computer vision have focused on excellent performance with diverse solutions. 


Existing conditional NFs for inverse problems in imaging occasionally generate unintended erroneous image samples. In super-resolution space generation, similar artifacts were observed in multiple independent works~\cite{lugmayr2021ntire,lugmayr2022ntire}. For example, Song \textit{et al.} reported that those artifacts occurred for more than 2\% of all test images~\cite{Song_2022_CVPR} and we confirmed that these artifacts occasionally appear as illustrated in the top row of Figure~\ref{fig:abstract_fig}.
Unintended artifacts were also observed in another computer vision task with conditional NFs. 
In low-light image enhancement~\cite{wang2021low}, we also revealed that black 
regions with $\mathtt{Inf}$ values sometimes occur for certain conditional inputs as we sample diverse images as in the bottom row of Figure~\ref{fig:abstract_fig}.
In \emph{unconditional} NFs, such as Glow~\cite{kingma2018glow}, similar artifacts called ``exploding inverse" were observed~\cite{pmlr-v130-behrmann21a}, which were known to occasionally occur only when the training and test sets come from different distributions (\textit{e.g.}, training with CIFAR-10~\cite{krizhevsky2009learning}, testing with tinyImageNet~\cite{xiao2019learning}). However, in conditional NFs for inverse problems in imaging, artifacts may sometimes occur even when the training and test sets follow the same distribution, suggesting that the existing ``exploding inverse'' is insufficient to explain this phenomenon.

In this work, we address the robustness issue for solving inverse problems in imaging using conditional NFs by investigating the origins of these artifacts and proposing how to avoid them.
Firstly, we empirically and theoretically reveal that artifacts arising from conditional NFs for inverse problems are caused by a mechanism very similar to that of unconditional NFs' exploding inverses~\cite{pmlr-v130-behrmann21a}. This implies that although the conditional inputs that yielded exploding inverses are sampled from the same distribution as the training dataset, they may be out-of-distribution (OOD) data from the perspective of the conditioning network. We then validate this remark (Remark \ref{remark-OOD}) by showing that the probability of causing erroneous pixels 
is highly correlated with a Mahalanobis distance-based OOD score~\cite{lee2018simple} for inverse problems. Lastly, based on our investigations, we propose another remark (Remark \ref{remark1}) on how to avoid the exploding inverses in conditional NFs for inverse problems in imaging. As a simple remedy to meet the criteria of our remark, we suggest substituting the affine coupling layers with the modified rational-quadratic (RQ) spline coupling layers~\cite{durkan2019neural} in NFs, to encourage the robustness of generated image samples. Our experimental results demonstrated that our suggested methods effectively suppressed exploding inverses often occurring in conditional NFs for super-resolution space generation and low-light image enhancement.
The contributions of this paper are summarized as follows:
\begin{itemize}
    \item Revealing theoretically and experimentally that exploding inverses also occur in conditional affine coupling flows for inverse problems occasionally, even when the training and test dataset are sampled from the same distribution.
\item Investigating that conditional inputs for yielding exploding inverses are out-of-distribution, from the perspective of the conditioning network, for normalizing flows (NFs) of inverse problems in imaging.
    \item Proposing a remark on how to avoid the exploding inverses in conditional NFs and demonstrating how to use it by considering other factors such as performance.
    \item Demonstrating that the proposed method effectively suppressed erroneous samples in 
    2D toy experiment, super-resolution space generation and low-light image enhancement.
\end{itemize}

\section{Preliminaries}
\label{sec:rel}
\subsection{Conditional normalizing flow}\label{sec:2.1}
NFs~\cite{rezende2015variational,papamakarios2021normalizing} learn a probability distribution from a dataset and can be used as both samplers and density estimators. 
Let $\mathcal{D}$ be a dataset from the true target probability distribution $p_{\mathbf{x}}$. 
One can utilize NF by fitting a flow-based model $q_\mathbf{x}$ to the true target distribution $p_{\mathbf{x}}$ using a simple base probability distribution $q_{\mathbf{z}}$ (\textit{e.g.}, standard normal distribution) and a diffeomorphic (\textit{i.e.}, invertible and differentiable) mapping $f_{\boldsymbol{\theta}}:\mathcal{X} \rightarrow \mathcal{Z}$ where $\mathcal{X}$ and $\mathcal{Z}$ are compact subsets of $\mathbb{R}^D$ with the following density transformation: 
\begin{equation}\label{eq1.1}
    q_{\mathbf{x}}(\mathbf{x}) = q_{\mathbf{z}}(f_{\boldsymbol\theta}(\mathbf{x})) \left\lvert \mathrm{det} \frac{\partial f_{\boldsymbol\theta}}{\partial \mathbf{x}} (\mathbf{x}) \right\rvert.
\end{equation} 
For $\mathcal{D}=\{\mathbf{x}^{(n)}\}_{n=1}^{N}$ where 
$\mathbf{x}^{(1)}, \dots, \mathbf{x}^{(N)}$ are 
samples from $p_{\mathbf{x}}$, 
NFs are 
trained on $\mathcal{D}$ by minimizing the following negative log-likelihood (NLL):
\begin{equation}\label{eq1.2}
    \mathcal{L}_{\text{NLL}} = -\frac{1}{N}\sum_{n=1}^{N} \log q_\mathbf{x} (\mathbf{x}^{(n)})
    \xrightarrow{ N \to \infty } -\mathbb{E}_{\mathbf{x} \sim p_{\mathbf{x}}}[\log q_{\mathbf{x}}(\mathbf{x})].
\end{equation}
Conditional NFs can be defined by simply changing the network in \eqref{eq1.1} to be conditional to $\mathbf{y}$ so that
\begin{equation}\label{eq1.3}
    q_{\mathbf{x}|\mathbf{y}}(\mathbf{x}|\mathbf{y}) = q_{\mathbf{z}}(f_{\boldsymbol\theta}(\mathbf{x};\mathbf{y})) \left\lvert \mathrm{det} \frac{\partial f_{\boldsymbol\theta}}{\partial \mathbf{x}} (\mathbf{x};\mathbf{y}) \right\rvert.
\end{equation}
For inverse problems in imaging, \eqref{eq1.3} is equivalent to modeling the posterior distribution $p_{\mathbf{x}|\mathbf{y}}(\mathbf{x}|\mathbf{y})$ where $\mathbf{y}$ is a corrupted measurement from a clean image $\mathbf{x}$.

\subsection{Coupling transformations}\label{sec:rel_1}
For a given $\mathbf{y}$, conditional NFs 
can obtain multiple possible $\mathbf{x}$ 
when used as samplers with different $\mathbf{z} \sim q_{\mathbf{z}}$ thanks to the one-to-one relationship between $\mathbf{z}$ and $\mathbf{x}$. However, ensuring this one-to-one relationship has limited 
network structures.
In order for NFs to be efficiently trainable with the NLL \eqref{eq1.2}, $f_{\boldsymbol{\theta}}$ must not only be invertible, but also have a tractable Jacobian determinant. Although many successful deep learning networks in the image domain employ 3$\times$3 convolution, max-pooling and ReLU (Rectified Linear Unit) layers, NFs cannot employ them since such layers are not invertible. 
A number of studies have found that only a few layers are appropriate for NFs in the image domain~\cite{DBLP:journals/corr/DinhKB14,rezende2015variational,DBLP:conf/iclr/DinhSB17,papamakarios2017masked,kingma2016improved,kingma2018glow}.

The conditional coupling layers are what make the NF ``conditional''; hence they are frequently used as the main layer.
A conditional coupling transformation~\cite{DBLP:journals/corr/DinhKB14, DBLP:conf/iclr/DinhSB17} $\phi : \Omega \rightarrow \Omega \subseteq \mathbb{R}^D$ is defined as
\begin{equation}\label{eq2.1.1}
    \phi(\mathbf{x})_i = \begin{cases}
    c(x_i;\mathbf{h}_i) & \text{for } i = d, \dots, D, \\
    x_i & \text{for } i = 1, \dots, d-1,
    \end{cases}
\end{equation}
where $\mathbf{h}_i = \mathrm{NN}(x_{1:d-1},g_{\boldsymbol{\theta}}(\mathbf{y}))$, $\mathrm{NN}$ is an arbitrary neural network, $g_{\boldsymbol{\theta}}$ is an encoder for the conditional input $\mathbf{y}$, and $c(\cdot;\mathbf{h}(\mathbf{y})):\Omega' \rightarrow \Omega' \subseteq \mathbb{R}$ is an invertible function parameterized by a vector $\mathbf{h}$.
The Jacobian determinant of this transformation is easily obtained from the derivative of $c$, expressed as $\mathrm{det} \left( {\partial \phi}/{\partial \mathbf{x}} \right) = \Pi_{i=d}^{D} {\partial c(x_i;\mathbf{h}_i)}/{\partial x_i}$.
The inverse of $\phi$ is obtained as
\begin{equation}\label{eq2.1.2}
    \phi^{-1}(\mathbf{x})_i = \begin{cases}
    c^{-1}(x_i;\mathbf{h}_i) & \text{for } i = d, \dots, D, \\
    x_i & \text{for } i = 1, \dots, d-1.
    \end{cases}
\end{equation}

Many works employ affine transformations as $c$:
\begin{equation}\label{eqaffine}
    c(x_i;\mathbf{h}_i(\mathbf{y}))=s_i(\mathbf{y}) x_i + t_i(\mathbf{y})
\end{equation}
where $\mathbf{h}_i = (s_i, t_i)$ due to computational efficiency for their Jacobian and inverse as well as sufficient expressive power~\cite{DBLP:journals/corr/KingmaW13, DBLP:conf/iclr/DinhSB17}.
Thus, affine coupling transformations are suitable for generating images~\cite{kingma2018glow,ho2019flow++,lugmayr2020srflow,sukthanker2022generative} and speeches~\cite{prenger2019waveglow,DBLP:conf/aaai/HeZ0LHY22} with 
large dimensions $D$. 

There are also other coupling transformations for conditional NFs such as splines~\cite{muller2019neural, durkan2019cubic, dolatabadi2020invertible, durkan2019neural, rezende2020normalizing} or sigmoids~\cite{kohler2021smooth}, which are more complex than affine coupling transformations. Splines are diffeomorphic piecewise polynomials or rational functions. 
Even though employing conditional spline-based coupling layers is possible, splines or sigmoids are computationally challenging over 
affine transformations. Thus, they were not as popular as affine coupling transformations for imaging applications, 
but rather prominent in modeling probability distributions of smaller dimensions such as molecular structures~\cite{wu2020stochastic, kohler2021smooth}. 


\subsection{Conditional NFs for inverse problems}\label{sec:rel_2}

In most conditional NFs, the affine coupling transformations and affine injectors are the only components that depend on $\mathbf{y}$. NFs with this structure have successfully solved various inverse problems in imaging~\cite{lugmayr2020srflow,Song_2022_CVPR,wang2021low,kim2021noise,wang2022learning}. 

SRFlow~\cite{lugmayr2020srflow} has achieved excellent performance on super-resolution space generation by adapting Glow~\cite{kingma2018glow} 
as its backbone. It has also been extended to 
many variants such as
\cite{jo2021srflow,liang2021hierarchical,kim2021noise,lugmayr2022normalizing,Song_2022_CVPR}.
In low-light image enhancement, LLFlow~\cite{wang2021low} and TSFlow~\cite{wang2022learning} have successfully utilized conditional NF to reconstruct normally exposed images from low-quality inputs.
Some studies dealt with other inverse problems such as inpainting~\cite{lu2020structured}, dehazing~\cite{wu2022bin}, denoising~\cite{lu2020structured} and colorization~\cite{ardizzone2019guided}. 



\section{On the Robustness of Conditional NFs}


We revisited the work of Behrmann \textit{et al.}~\cite{pmlr-v130-behrmann21a} for exploding inverses in unconditional NFs and described the clear differences between that and our work on exploding inverses in conditional NFs. With a simple toy example, we identified that exploding inverses can also occur in the conditional NF using affine coupling layer, if the conditional input is OOD.
Then, we analyzed that the exploding inverse can be induced in the full conditional NF models, when the conditional inputs are OOD from the perspective of the conditional input encoder, even though they are in-distribution in human's perspective 
(\textit{i.e.}, $g_{\boldsymbol{\theta}}(\mathbf{y})$ is OOD even though $\mathbf{y}$ is in-distribution). We further investigated this phenomenon for two concrete examples: super-resolution space generation (FS-NCSR~\cite{Song_2022_CVPR}) and low-light image enhancement (LLFlow~\cite{wang2021low}). Lastly, we elaborate the conditions on how to avoid the exploding inverse for inverse problems in imaging and suggest a remedy to meet all the criteria.

\subsection{Exploding inverses in unconditional NFs}
Behrmann \textit{et al.}~\cite{pmlr-v130-behrmann21a} discovered and named ``exploding inverse'' in unconditional NFs. We revisit this work and reorganize the relevant parts of their work as follows. 

\begin{proposition}{\textbf{(Exploding inverses in unconditional NFs)}} 
If $f_\theta : \mathcal{X} \rightarrow \mathcal{Z} \subseteq \mathbb{R}^D$ is an unconditional NF using the affine coupling transformation and trained with a dataset from the distribution $p_\mathbf{x}$, then there exist many $\mathbf{x} \not \sim p_\mathbf{x}$ s.t. $\lVert \mathbf{x} \rVert_\infty \ll \lVert f^{-1}_\theta (f_\theta(\mathbf{x})) \rVert_\infty$.
\end{proposition}
Since $\mathbf{x} \not \sim p_\mathbf{x}$ suggests $f_\theta(\mathbf{x}) \not \sim q_\mathbf{z}$, it is reasonable to say 
that errors can occur.
The problem we address in this work is clearly different. It can be summarized as follows.
\begin{proposition}{\textbf{(Artifacts in conditional NFs)}} 
If $f_\theta : \mathcal{X} \times \mathcal{Y} \rightarrow \mathcal{Z} \subseteq \mathbb{R}^D$ is a conditional NF using the conditional affine coupling transformation and trained with a dataset from the distribution $p_{\mathbf{x}|\mathbf{y}}$, then there exist many $\mathbf{y} \sim p_\mathbf{y}$, $\mathbf{z} \sim q_\mathbf{z}$  such that $f^{-1}_\theta (\mathbf{z};\mathbf{y})$ is erroneous. 
\end{proposition}
The most significant difference between 
these two propositions is that the sample (\textit{i.e.}, $f^{-1}_\theta (\mathbf{z};\mathbf{y})$) can be erroneous even though the inputs (\textit{i.e.}, $\mathbf{y}$ and $\mathbf{z}$) are in-distribution in human's perspective. 
To help understanding where these errors come from, we build and verify this proposition, and then investigate which $\mathbf{y}$ generates artifacts in the next subsections.

\begin{figure}
    \centering
\begin{tabular}{lcc}
\begin{tabular}[c]{@{}l@{}}\small Training\\ \small data\end{tabular}                                                      & \multicolumn{2}{c}{\adjustbox{valign=m}{{\includegraphics[width=0.3\columnwidth ]{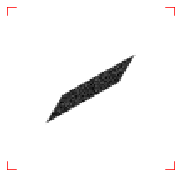} }}} \\
& \small Affine  & \small Modified RQ-spline (ours) \\
\begin{tabular}[c]{@{}l@{}}\small Flow\\ \small samples\end{tabular}             &   \adjustbox{valign=m}{{\includegraphics[width=0.3\columnwidth ]{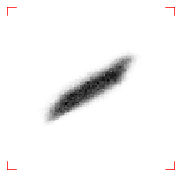} }}      &       \adjustbox{valign=m}{{\includegraphics[width=0.3\columnwidth ]{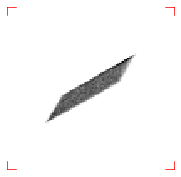} }}     \\
\begin{tabular}[c]{@{}l@{}}\small Flow \\ \small samples \\ \small for OOD\end{tabular} &   \adjustbox{valign=m}{{\includegraphics[width=0.3\columnwidth ]{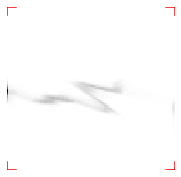} }}      &     \adjustbox{valign=m}{{\includegraphics[width=0.3\columnwidth ]{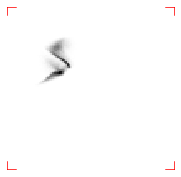} }}      
\end{tabular}
    \caption{2D toy experiment results. The first row shows the training data (uniformly distributed). The second and third row show the flow samples for in-distribution/OOD conditional input (\textit{i.e.}, $\mathbf{y}_{\text{in}}$ and $\mathbf{y}_{\text{OOD}}$), respectively. The left and right columns show the results of employing the conditional affine/RQ-spline coupling layers, respectively. The displayed area is $[-1,1]^2$, marked with red angle brackets.}
    \label{fig:2dToy}
\end{figure}

\subsection{Exploding inverse in conditional NFs}\label{sec:3.1}

\subsubsection{A 2D toy experiment}
\label{sec:2dToy}
We constructed a simple 2D toy experiment, 
demonstrating that the conditional affine coupling flows can suffer from exploding inverse 
for OOD conditional inputs.
A forward model of the inverse problem is selected as follows:
\begin{equation}
    \mathbf{y}_{\text{in}} = \mathbf{Ax} + \boldsymbol{\epsilon}, \mathbf{A} = \left[
\begin{small}
\begin{matrix}
    0.7 & 0.3 \\
    0.3 & 0.7 \\
\end{matrix}
\end{small}
\right],
\boldsymbol{\epsilon} \sim \mathcal{N}(0,\sigma_n^2 \mathbf{I})
\end{equation}
where $\sigma_n = 0.01$, $\mathbf{x},\mathbf{y}_{\text{in}} \in \mathbb{R}^2$.
Training data was generated as illustrated in the first row of Figure~\ref{fig:2dToy}. 
We also generated OOD conditional input $\mathbf{y}_{\text{OOD}}$ by shifting $\mathbf{y}_{\text{in}}$ as $
    \mathbf{y}_{\text{OOD}}  = \mathbf{y}_{\text{in}} + \left[
\begin{matrix}
    0.8 & -0.8 
\end{matrix}
\right]^\mathrm{T}.$
See the supplementary material for further information on the toy experiment. 

\begin{figure}
    \centering
    \begin{subfigure}{0.65\columnwidth}
        \includegraphics[width=\textwidth]{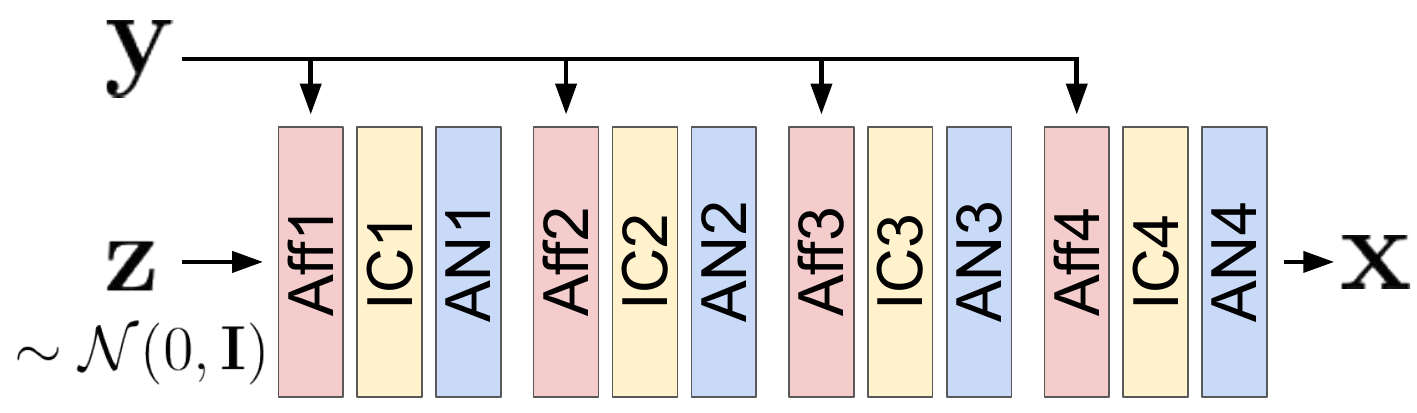}
        \caption{Network architecture}
        \label{fig:3.1}
    \end{subfigure}
    \begin{subfigure}{0.8\columnwidth}
        \includegraphics[width=\textwidth]{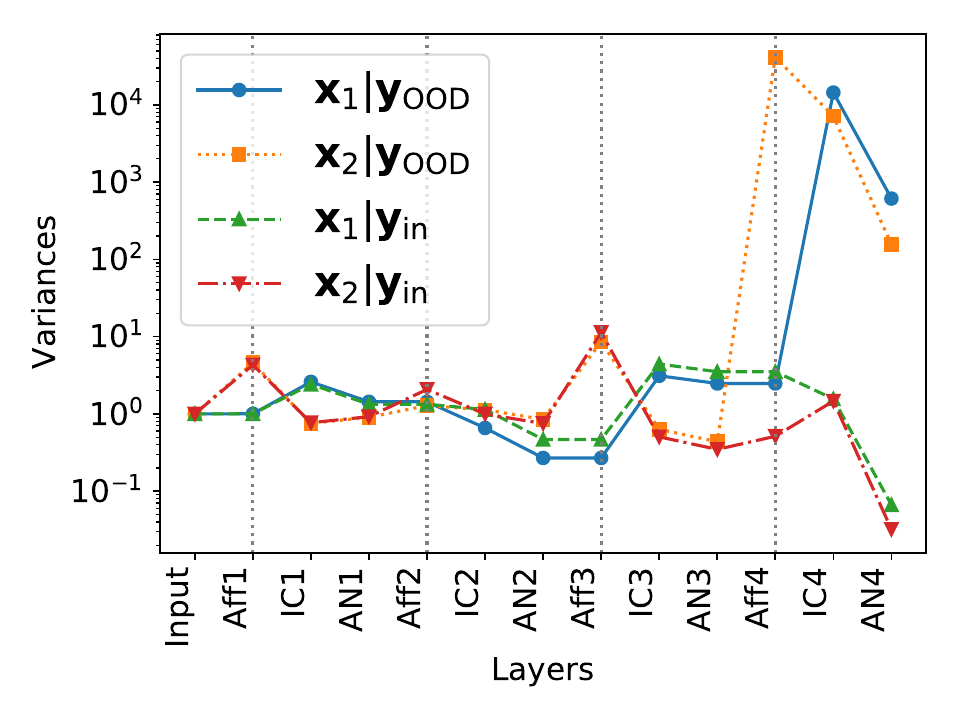} 
   \vspace{-2em}
        \caption{Variances of features}
        \label{fig:3.2}
    \end{subfigure}
   \vspace{-.5em}
    \caption{(a) Network architecture for toy experiment. (b) Variances of features for in-distribution and OOD conditional inputs. \textsf{Aff}, \textsf{IC} and \textsf{AN} denote the conditional affine coupling, invertible 1$\times$1 convolution and activation normalization layers, respectively.}
    \label{fig:2dToyResults}
\end{figure}

In the left column of Figure~\ref{fig:2dToy}, flow samples for in-distribution (\textit{i.e.}, $\mathbf{y}_{\text{in}}$) show that the flow model in Figure~\ref{fig:3.1} learned the distribution well, but 
flow samples for OOD (\textit{i.e.}, $\mathbf{y}_{\text{OOD}}$) show that the flow model failed to learn the distribution correctly.
Although the support of the distribution of $\mathbf{x}$ was a subset of a small region (\textit{i.e.}, $\mathrm{supp}(p_{\mathbf{x}|\mathbf{y}}(\mathbf{x}|\mathbf{y}_{\text{in}})) \subset [-1,1]^2$), the flow model generated samples that were located outside the region (\textit{i.e.}, $\mathrm{supp}(q_{\mathbf{x}|\mathbf{y}}(\mathbf{x}|\mathbf{y}_{\text{OOD}})) \not\subset [-1,1]^2$). Note that we put all samples outside the region $[-1,1]^2$ on the edges. 
This corresponds to clipping the pixel value of image samples to $[0,1]$ (for unsigned 8-bit integer, $[0, 255]$), which can explain the saturated color compositions of the artifacts that have been observed in conditional NF-based methods for super-resolution space generation and low-light image enhancement
as illustrated in the right column of Figure~\ref{fig:abstract_fig}.

We further investigated this instability problem by looking into 
the variances of features (Figure~\ref{fig:3.2}) in each layer of the flow model (Figure~\ref{fig:3.1}) where the conditional inputs are either in-distribution or OOD. 
Before the 4th affine coupling layer (\textsf{Aff4}), features have very similar variances at each layer. However, at \textsf{Aff4}, the variance explodes more than 10,000 times for the OOD case, whereas the variance is maintained for the in-distribution case.
Even though our setting was different from unconditional NFs, the cause and the results are very similar to ``exploding inverse''~\cite{pmlr-v130-behrmann21a}.
Thus, these results support the following remark:
\begin{remark}\label{remark-OOD}
Conditional NFs with affine coupling layers can generate erroneous samples due to exploding inverse for certain OOD conditional input.
\end{remark}
\noindent Section~\ref{sec:theory} presents a theoretical analysis with a simplified model for exploding inverse to support Remark~\ref{remark-OOD}. Section~\ref{sec:OOD} verifies that Remark~\ref{remark-OOD} is valid in full-size networks for real inverse problems in imaging.

\subsubsection{Theoretical analysis on exploding inverse}\label{sec:theory}

From the convex optimization perspective, we explain why the exploding inverse occurs for certain 
conditional inputs.

As in \eqref{eq1.2}, $f_{\boldsymbol{\theta}}$ is trained by minimizing $\mathcal{L}_\mathrm{NLL} = -\mathbb{E}_{\mathbf{x},\mathbf{y} \sim \hat{p}_{\mathbf{x},\mathbf{y}}}[\log q_{\mathbf{x}|\mathbf{y}}(\mathbf{x}|\mathbf{y})]$.
For 
simplicity, we assume that a model $f_{\boldsymbol{\theta}}$ consists of one conditional affine coupling layer where $\mathbf{x},\mathbf{y} \in \mathbb{R}^2$.
Using \eqref{eq1.3} with \eqref{eq2.1.1} and \eqref{eqaffine} when $d=D=2$, we obtain the following NLL loss $\mathcal{L}_\mathrm{NLL}$
\begin{equation}\label{eq3.1.2.1}
\begin{aligned}
    & = -\mathbb{E}_{\mathbf{x},\mathbf{y}}\left[\log q_{\mathbf{z}}(f_{\boldsymbol\theta}(\mathbf{x};\mathbf{y})) + \log \left\lvert \mathrm{det} \frac{\partial f_{\boldsymbol\theta}}{\partial \mathbf{x}} (\mathbf{x};\mathbf{y}) \right\rvert \right]\\
    & = \mathbb{E}_{\mathbf{x},\mathbf{y}}\left[ 
    \frac{\lVert f_{\boldsymbol\theta}(\mathbf{x};\mathbf{y}) \rVert_2^2}{2\sigma_z^2} - \log \left\lvert \mathrm{det} \left[
\begin{matrix}
    1 & 0 \\
    * & s_1 \\
\end{matrix}
\right]
\right\rvert \right]\\
& = \mathbb{E}_{\mathbf{x},\mathbf{y}}\left[ 
\frac{x_1^2 + (s_1 x_1 + t_1)^2}{2\sigma_z^2} -\log(s_1)\right],
\end{aligned}
\end{equation}
where $\mathbf{z}$ is assumed to be Gaussian 
and $(s_1, t_1)$ are the functions of $\mathbf{y}$. Thus, 
\eqref{eq3.1.2.1} is a convex function of $(s_1,t_1)$.
This NLL loss is unbounded below, 
so that there is a degenerative case for $(s_1, t_1)$, \textit{i.e.}, $s_1 \rightarrow \infty$ with $t_1 \rightarrow -s_1 x_1$. Similarly, Kirichenko~\textit{et al.}~\cite{kirichenko2020normalizing} reported that $s$ often diverges to infinity in the affine coupling layer, which led to performance degradation in density estimation.

This undesirable unboundedness of the NLL loss can be avoided by setting an upper bound on $s_1$ so that the optimization problem becomes proper 
(\textit{i.e.}, bounded below): 
\begin{equation}
    \min_{0 < s_1 \leq 1,t_1} \frac{x_1^2 + (s_1 x_1 + t_1)^2}{2\sigma_z^2} -\log(s_1), \label{eq3.1.2.2}
\end{equation}
which has the analytic solution $(s_1, t_1) = (1, -x_1)$.
Interestingly, recent flow models such as SRFlow~\cite{lugmayr2020srflow} and LLFlow~\cite{wang2021low} set an upper bound of $s_i$ to $1$ without any theoretical discussion like 
our analysis with \eqref{eq3.1.2.2}.

The exploding inverse that was observed in Figure~\ref{fig:2dToyResults} can be explained theoretically with the convex optimization \eqref{eq3.1.2.2}.
For the in-distribution conditional input $\mathbf{y}$, $(s_1(\mathbf{y}), t_1(\mathbf{y}))$ will usually yield values close to the optimal point $(1, -x_1)$. 
Since $s_1$ is close to $1$, the conditional affine coupling layer will not 
increase the variance of features much. However, for the OOD conditional input, 
there may be some $(s_1(\mathbf{y}), t_1(\mathbf{y}))$ that are far from the optimal point $(1, -x_1)$, which would 
result in $s_1 \ll 1$. Considering the sampling process, which is the reverse process of density estimation as in \eqref{eq2.1.2}, this can significantly increase the variance of features since $1/s_1 \gg 1$, which causes exploding inverse and thus generates erroneous images.
One may think that setting a proper lower bound on $s_1$ could resolve 
this issue (\textit{e.g.}, $0.1<s_1\leq 1.1$ in \eqref{eq3.1.2.2}). 
Section~\ref{sec:experiments} provides the experimental results that this na\"ive approach cannot solve it. 

\subsection{OOD conditional inputs for conditional NFs}\label{sec:OOD}

Here we verify that certain conditional inputs that cause errors are OOD, even though they are in-distribution in Human's eye. Specifically, we check the difference between $\mathbf{y}_\text{in}$ and $\mathbf{y}_\text{OOD}$ by investigating 
the encoder output $g_{\boldsymbol{\theta}}$ for the conditional input $\mathbf{y}$.
The Mahalanobis distance~\cite{lee2018simple} was selected to measure these differences. 
The Mahalanobis distance of a point $\mathbf{v}$ from a probability measure $p$ is defined as
\begin{equation}\label{eq3.2.1}
    d_M(\mathbf{v},p) = \sqrt{(\mathbf{v}-\mu_p)^\mathrm{T} \Sigma_p^{-1} (\mathbf{v}-\mu_p)},
\end{equation}
where $\mu_p = \mathbb{E}_{\mathbf{u}\sim p}[\mathbf{u}]$, $\Sigma_p = \mathbb{E}_{\mathbf{u}\sim p}[\mathbf{u}\mathbf{u}^\mathrm{T}]$.
To check the difference from the perspective of $g_{\boldsymbol{\theta}}$ rather than the data themselves, we compare  $d_M(g_{\boldsymbol{\theta}}(\mathbf{y}_{\text{in}}), g_{\boldsymbol{\theta}\#}\hat{p}_{\mathbf{y}})$ and $d_M(g_{\boldsymbol{\theta}}(\mathbf{y}_\text{OOD}), g_{\boldsymbol{\theta}\#}\hat{p}_{\mathbf{y}})$, where $g_{\boldsymbol{\theta}\#}\hat{p}_{\mathbf{y}}$ is a pushforward of $\hat{p}_{\mathbf{y}} = (1/N)\sum_{j=1}^{N} \delta_{\mathbf{y}^{(j)}}$ with respect to $g_{\boldsymbol{\theta}}$. For simplicity, we denote our OOD score as follows:
\begin{equation}\label{eq3.2.2}
    s_{\text{OOD}}(\mathbf{y'}) = d_M(g_{\boldsymbol{\theta}}(\mathbf{y'}), g_{\boldsymbol{\theta}\#}\hat{p}_{\mathbf{y}}).
\end{equation}
We calculate $s_{\text{OOD}}$ where $\mathbf{y'}$ is a cropped patch in the test set and $\hat{p}_{\mathbf{y}}$ is the distribution of the training set. 

Figure~\ref{fig:abstract_fig} shows erroneous samples generated from the conditional inputs with the highest OOD score in the test set. 
Both conditional inputs generated erroneous samples. To further validate that conditional inputs with high OOD score are prone to generate exploding inverses, we plotted the probability of pixel errors versus the OOD score in Figure \ref{fig:4}. See the supplementary material for more details.
Figure \ref{fig:4a} shows that the top 300 ranked patches among the DIV2K validation set (\textit{i.e.}, test set from the same distribution as the training set) generally have higher probability of generating erroneous pixels compared to the dashed horizontal line, which denotes the average error probability for all patches. 
To investigate severe OOD cases, we utilized the Enhanced Urban100 (EUrban100) dataset~\cite{huang2015single}, which can be perceived as a prominent example of severe OOD to the human eye.
The logistic regression result shows that conditional inputs with high OOD score are prone to generate pixel errors. To summarize, conditional inputs which frequently generates erroneous images are OOD in the perspective of the conditioning network $g_{\boldsymbol{\theta}}$.

\begin{figure}
    \centering
    \begin{subfigure}{\columnwidth}
        \centering
        \includegraphics[width=0.99\textwidth]{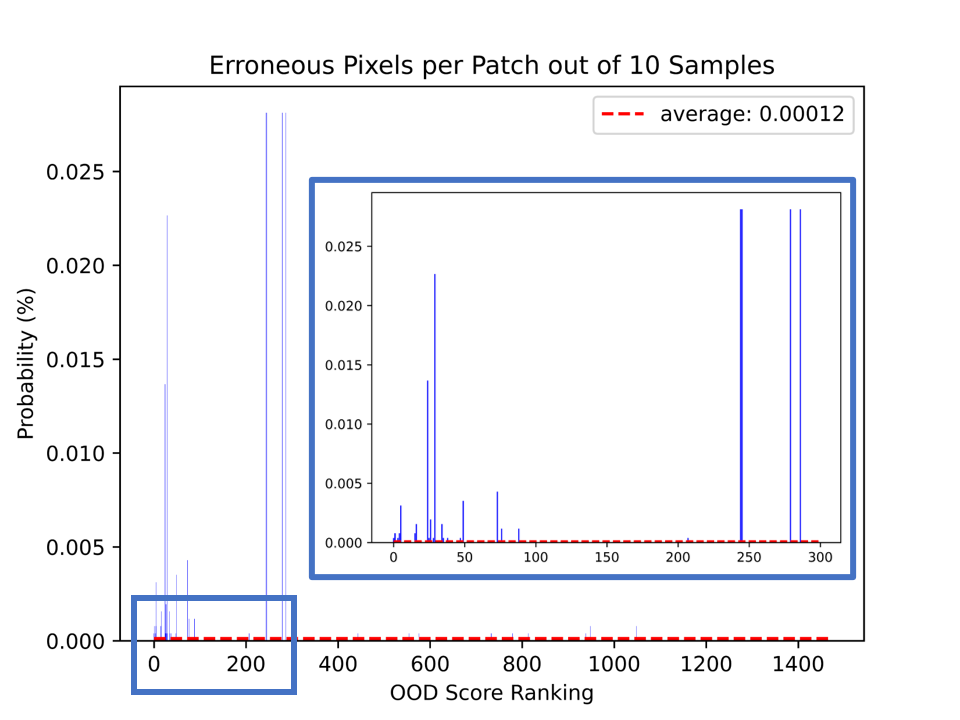}
        \vspace{-0.5em}
        \caption{Error pixel rate sorted by $s_{\text{OOD}}$ ranking}
        \label{fig:4a}
    \end{subfigure}
    \begin{subfigure}{\columnwidth}
        \centering
        \includegraphics[width=0.7\textwidth]{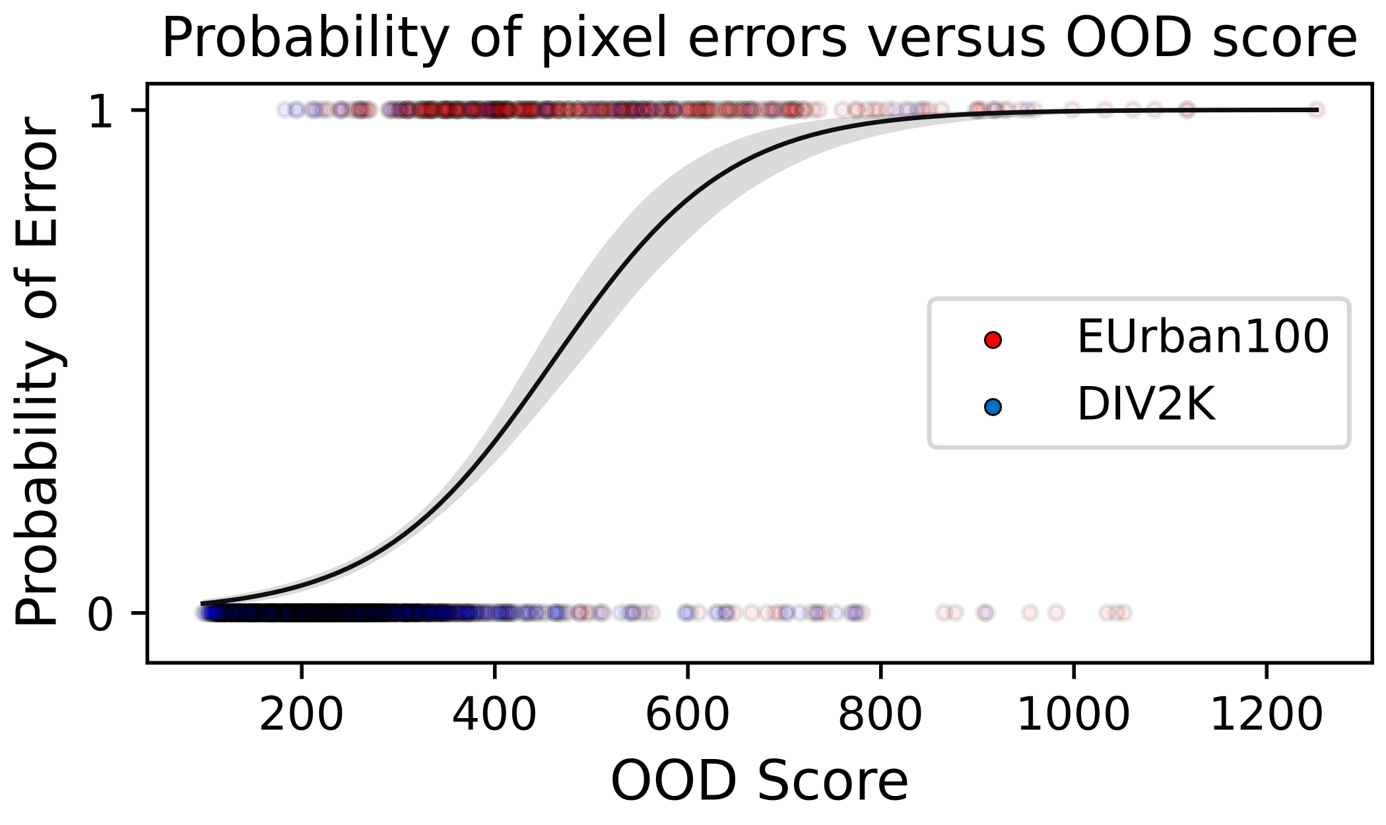}
        \vspace{-0.5em}
        \caption{Logistic regression: $\exists$ Error (Pixel) vs. $s_{\text{OOD}}$}
        \label{fig:4b}
    \end{subfigure}
    \caption{(a) Erroneous pixels for the patches (\textit{i.e.}, conditional inputs) ranked with their OOD scores ($s_{\text{OOD}}$) and their average (dashed horizontal line). (b) Logistic regression: the probability of existence of pixel error versus $s_{\text{OOD}}$.}
    \label{fig:4}
        \vspace{-.5em}
\end{figure}

\subsection{On how to avoid exploding inverse}\label{sec:rqspline}

From the experimental and theoretical investigations in the previous subsections~\ref{sec:3.1} and \ref{sec:OOD}, we propose a remark on how to avoid exploding inverse.
For a diffeomorphic function of a conditional coupling transformation $c: \mathbb{R} \rightarrow \mathbb{R}$,
let $c'(x) \in [s_l,s_u]$ (\textit{i.e.}, $c$ is bi-Lipschitz continuous). By generalizing the optimization problem for the affine transformation to this function $c$, the same phenomenon as that in Section~\ref{sec:theory} can be observed:
\begin{equation}
    c'(x) \begin{cases}
    \simeq s_u & \text{for in-distribution } \mathbf{y},\\
    \ll s_u & \text{for OOD } \mathbf{y}.
    \end{cases}
\end{equation}
To avoid exploding inverse, coupling transformations must satisfy the following remark:
\begin{remark}\label{remark1}
    To avoid exploding inverse, the derivatives of the element-wise transformation $c$ of the conditional coupling layer must yield similar lower and upper bounds when the input has a sufficiently large absolute value. In other words, $c'(x)\simeq s_u$ for $\mathbf{y}$ is OOD and $|x| \gg 1$.
\end{remark}

While Remark~\ref{remark1} can guide one to design or to select a proper coupling transformation for conditional NFs to avoid exploding inverse, there are also other conditions to consider for performance such as sufficient expressive power and computational efficiency.
In this work, we demonstrate how to select a coupling transformation considering both Remark~\ref{remark1} and other conditions like performance among existing ones. The same rules can be utilized for designing a new one.

\paragraph{A solution to satisfy Remark~\ref{remark1}:} We can set $c (x) = x + t$ for all $x \not\in (B_1,B_2)$, which satisfies Remark~\ref{remark1} ($c'(x)=1$ for all $x \in (-\infty,B_1) \cup (B_2,\infty)$) and computational efficiency (by using only a few parameters). However, this would not have sufficient expressive power. The additive coupling transformation~\cite{DBLP:journals/corr/DinhKB14} also lacks expressive power. In the meanwhile, spline-based transformations have better expressive power than affine transformations~\cite{muller2019neural, durkan2019cubic, dolatabadi2020invertible, durkan2019neural, rezende2020normalizing}, but with inefficient, relatively long computation.  

\begin{figure}
    \centering
    \begin{subfigure}[b]{0.35\columnwidth}
		\centering
		\includegraphics[width=\textwidth]{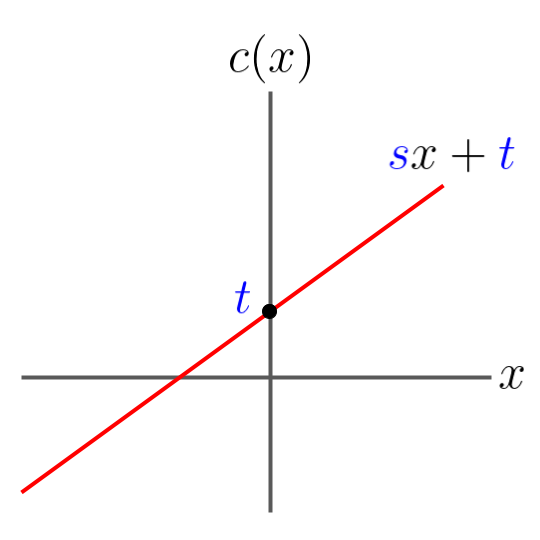}
		\caption{Affine}
		\label{fig:affine}
	\end{subfigure}
	\begin{subfigure}[b]{0.5\columnwidth}
		\centering
		\includegraphics[width=0.7\textwidth]{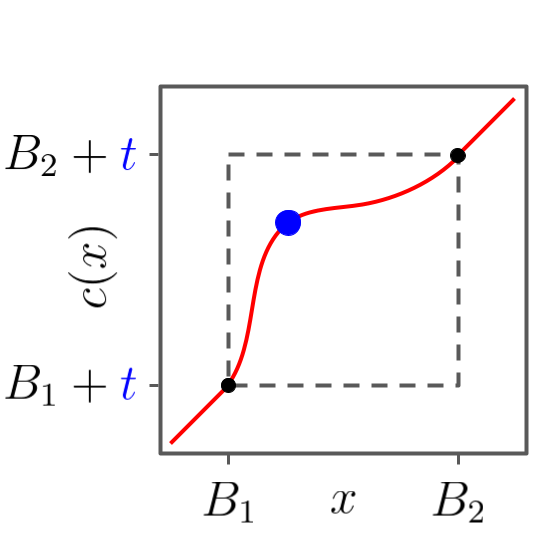}
		\caption{Modified RQ-spline (ours)}
		\label{fig:rqspline}
	\end{subfigure}
	\caption{(a) Affine and (b) the modified RQ-spline transformations for coupling layers. The learnable parameters are labeled in blue.}
	\label{transformations}
	        \vspace{-1em}
\end{figure}

RQ-spline coupling layer~\cite{durkan2019neural} satisfies Remark~\ref{remark1} and has sufficient expressive power. Figure~\ref{transformations} compares the affine and RQ-spline transformations. As in Figure~\ref{fig:rqspline}, we set only one out of three knots as learnable parameter to be computationally efficient. To maintain the expressive power with a small number of parameters, we propose to add a bias term $t$ (called the modified RQ-spline), which does not violate Remark~\ref{remark1}. Note that this method is an example which avoids exploding inverse while having reasonable computational efficiency and expressive power. Other choices or designs could be done for better performance, but our Remark~\ref{remark1} will be an important guideline to avoid potential errors due to exploding inverse. 

\section{Experimental Results}\label{sec:experiments}
\subsection{2D toy experiment}\label{sec:toy_result}
The right column in Figure \ref{fig:2dToy} shows the results of performing the 2D toy experiment in Section \ref{sec:2dToy} using the proposed modified RQ-spline coupling transformation instead of the affine coupling transformation. We also plotted variances of features (as Figure \ref{fig:2dToyResults}) in the supplementary material. Unlike the case with affine coupling transformation, the variance does not explode for OOD conditional inputs ($\mathbf{y}_{\text{OOD}}$) when RQ-spline coupling transformation is used. Therefore, all samples are included within the range shown in the figure (\textit{i.e.}, $\mathrm{supp}(q_{\mathbf{x}|\mathbf{y}}(\mathbf{x}|\mathbf{y}_{\text{OOD}})) \subset [-1,1]^2$).

\subsection{Super-resolution space generation}\label{sec:fsncsr}
\begin{figure*}
    \centering
    \begin{tabular}{ccccc}
        \includegraphics[width=0.175\textwidth]{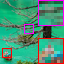} &
        \includegraphics[width=0.175\textwidth]{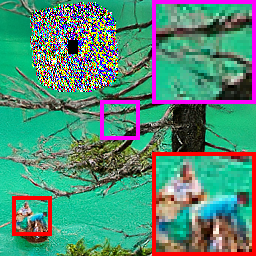} &
        \includegraphics[width=0.175\textwidth]{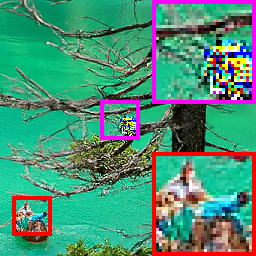} &
        \includegraphics[width=0.175\textwidth]{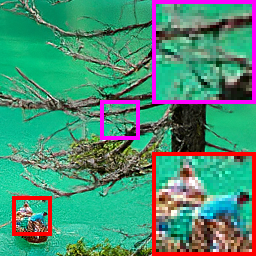} &
        \includegraphics[width=0.175\textwidth]{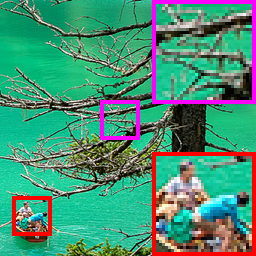} \\
        \includegraphics[width=0.175\textwidth]{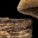} &
        \includegraphics[width=0.175\textwidth]{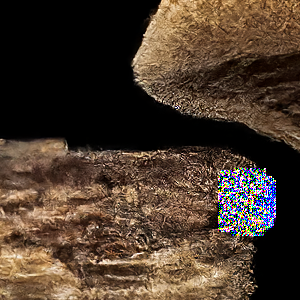} &
        \includegraphics[width=0.175\textwidth]{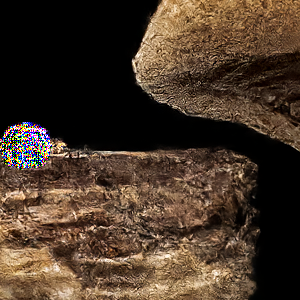} &
        \includegraphics[width=0.175\textwidth]{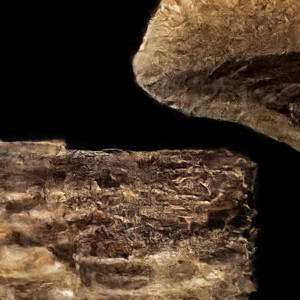} &
        \includegraphics[width=0.175\textwidth]{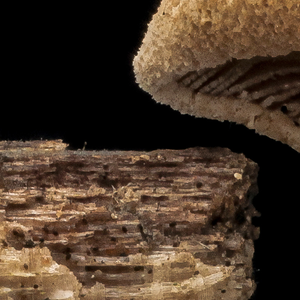} \\
        \includegraphics[width=0.175\textwidth]{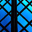} &
        \includegraphics[width=0.175\textwidth]{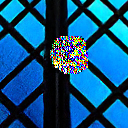} &
        \includegraphics[width=0.175\textwidth]{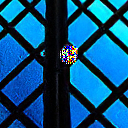} &
        \includegraphics[width=0.175\textwidth]{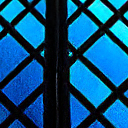} &
        \includegraphics[width=0.175\textwidth]{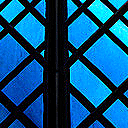} \\
        LR & FS-NCSR~\cite{Song_2022_CVPR} & FS-NCSR\textsuperscript{\textdagger} & Ours & Ground Truth\\
    \end{tabular}
               \vspace{-1em}
    \caption{Qualitative comparison of coupling transformation in super-resolution space generation. The first, second, and third row shows the samples from DIV2K~\cite{agustsson2017ntire} 4$\times$, DIV2K 8$\times$, and EUrban100 4$\times$. The {\textdagger} sign denotes that the lower bound of the scale parameter is $0.1$.}
    \label{tbl:table_of_figures}
\end{figure*}

We qualitatively and quantitatively compare generated samples from diverse datasets.
For some conditional inputs, FS-NCSR~\cite{Song_2022_CVPR} generated SR images with artifacts, as in the second column of Figure \ref{tbl:table_of_figures}.
Our model with the modified 
RQ-spline layers does not generate any erroneous image samples as illustrated in the fourth column of Figure \ref{tbl:table_of_figures}, even though the conditional inputs were severe OOD (EUrban100 4$\times$, whose average OOD score was about $2.15$ times larger than DIV2K 4$\times$ validation set).

For fair evaluation in detecting occasional errors, we use the average of the minimum and standard deviation of LR-PSNR among 10 samples, as LR-PSNR was one of the official evaluation metrics in the 2021, 2022 NTIRE challenges~\cite{lugmayr2021ntire,lugmayr2022ntire}.
Another metric $\% \mathtt{Inf}$, which was used in \cite{pmlr-v130-behrmann21a}, refers to the percentage of conditional inputs that generate at least one $\mathtt{Inf}$ pixel.
As shown in Table \ref{table:fsncsr}, 
 FS-NCSR\textsuperscript{\textdagger} (\textit{i.e.},  FS-NCSR with $0.1< s_1 \leq 1.1$ as discussed in Section \ref{sec:theory}) also suppressed $\mathtt{Inf}$ pixels. Still, erroneous images were sampled for the same conditional inputs, as shown in the third column of Figure \ref{tbl:table_of_figures}. In contrast, our method was completely free of errors and showed the best results. 


\begin{table*}[]
\centering
\small
\begin{tabular}{lccccccccc}
\toprule
             Train $\rightarrow$ Test              & \multicolumn{3}{c}{DF2K 4$\times$ $\rightarrow$ DIV2K 4$\times$}                    & \multicolumn{3}{c}{DF2K 8$\times$ $\rightarrow$ DIV2K 8$\times$}                    & \multicolumn{3}{c}{DF2K 4$\times$ $\rightarrow$ EUrban100 4$\times$ (OOD)} \\
           
\multicolumn{1}{l|}{Model} &  \multicolumn{1}{c}{$\% \mathtt{Inf}\downarrow$} & $\overline{\min}\uparrow$ & \multicolumn{1}{c|}{$\overline{\sigma}\downarrow$} &  \multicolumn{1}{c}{$\% \mathtt{Inf}\downarrow$} & $\overline{\min}\uparrow$ & \multicolumn{1}{c|}{$\overline{\sigma}$$\downarrow$} &  \multicolumn{1}{c}{$\% \mathtt{Inf}\downarrow$} & $\overline{\min}\uparrow$  & $\overline{\sigma}$$\downarrow$  \\ \midrule
FS-NCSR~\cite{Song_2022_CVPR}                       &  2  &   50.86      & \multicolumn{1}{c|}{0.202}      &   2   &   48.47        & \multicolumn{1}{c|}{0.461}      & 30 & 36.12           & 3.544       \\
FS-NCSR\textsuperscript{\textdagger}                    &  \textbf{0}  & 50.83          & \multicolumn{1}{c|}{0.077}      &   \textbf{0} & 49.50          & \multicolumn{1}{c|}{0.183}      &    \textbf{0} & 42.95           & 1.046       \\
Ours                       &     \textbf{0} &  \textbf{51.10}         & \multicolumn{1}{c|}{\textbf{0.012}}      &     \textbf{0} &   \textbf{50.20}        & \multicolumn{1}{c|}{\textbf{0.041}}      &    \textbf{0}  & \textbf{44.70}           & \textbf{0.136}       \\ \bottomrule
\end{tabular}
           \vspace{-.5em}
\caption{Quantitative comparison. The {\textdagger} sign denotes that the lower bound of the scale parameter is $0.1$.
`$\% \mathtt{Inf}$' refers to the percentage of conditional inputs that generate at least one $\mathtt{Inf}$ pixel out of 10 randomly generated latent codes, each with $\mathbf{z} \sim \mathcal{N}(\mathbf{0},\,\tau^2)$. $\overline{\min}$ and $\overline{\sigma}$ refer the average of the minimum and standard deviation of LR-PSNR, respectively. DF2K means the union set of DIV2K~\cite{agustsson2017ntire} and Flickr2K~\cite{timofte2017ntire}.}
\label{table:fsncsr}
 \vspace{-1em}
\end{table*}
\subsection{Low-light image enhancement}
\label{sec:llflow}

\begin{figure*}
    \centering
    \begin{tabular}{ccccc}
        \includegraphics[width=0.175\textwidth]{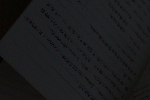} & 
        \includegraphics[width=0.175\textwidth]{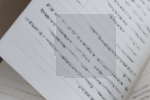} &
        \includegraphics[width=0.175\textwidth]{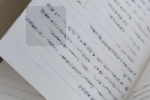} &
        \includegraphics[width=0.175\textwidth]{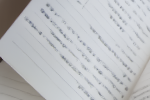} &
        \includegraphics[width=0.175\textwidth]{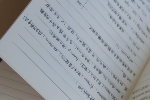} \\
        \includegraphics[width=0.175\textwidth]{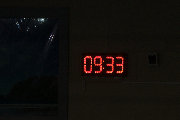} & 
        \includegraphics[width=0.175\textwidth]{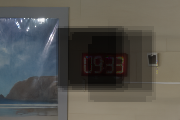} &
        \includegraphics[width=0.175\textwidth]{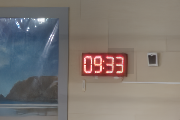} &
        \includegraphics[width=0.175\textwidth]{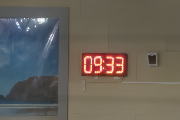} &
        \includegraphics[width=0.175\textwidth]{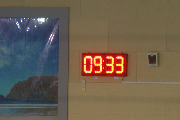} \\
        \includegraphics[width=0.175\textwidth]{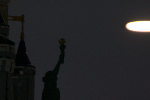} & 
        \includegraphics[width=0.175\textwidth]{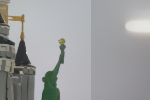} &
        \includegraphics[width=0.175\textwidth]{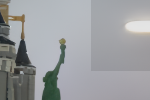} &
        \includegraphics[width=0.175\textwidth]{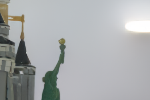} &
        \includegraphics[width=0.175\textwidth]{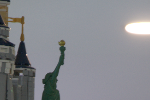} \\
        Input & 
        LLFlow~\cite{wang2021low} & LLFlow\textsuperscript{\textdagger} & Ours & Ground Truth\\
    \end{tabular}
               \vspace{-1em}
    \caption{Qualitative comparison of coupling transformation in low-light image enhancement. The top two rows and the bottom row are samples from the LOL~\cite{Chen2018Retinex} and VE-LOL~\cite{ll_benchmark} datasets, respectively. The {\textdagger} sign denotes that the lower bound of the scale parameter is $0.1$.}
    \label{fig:llflow}
\end{figure*}

We qualitatively compare the mean of 10 generated outputs of LLFlow~\cite{wang2021low} in Figure \ref{fig:llflow}. It is shown in the second column of Figure \ref{fig:llflow} that erroneous images (\textit{e.g.} black regions with $\mathtt{Inf}$ pixels around the clock) are generated through affine coupling layers while images generated through our method do not present artifacts, as in the fourth column of Figure \ref{fig:llflow}.

We also quantitatively compare the results in Table \ref{table:llflow}.
The first and second rows of Table \ref{table:llflow} show that the affine coupling transformation is prone to generating erroneous images whereas the modified RQ-spline coupling transformation is robust to OOD samples.
Even with the scale parameter of the affine transformation adjusted to exceed $0.1$, the black regions are still shown as in the third column of Figure \ref{fig:llflow}. See the supplementary material for details and more various erroneous images sampled from LLFlow.

\begin{table*}[]
\centering
\small
\begin{tabular}{lcccccccccc}
\toprule
                Train $\rightarrow$ Test    & \multicolumn{4}{c}{LOL $\rightarrow$ LOL}    
                           & \multicolumn{4}{c}{LOL $\rightarrow$ VE-LOL} \\
\multicolumn{1}{l|}{Model} & \begin{tabular}{@{}c@{}}$\% \mathtt{Inf}\downarrow$ \end{tabular} & PSNR$\uparrow$ & SSIM$\uparrow$ & \multicolumn{1}{c|}{LPIPS$\downarrow$} & \begin{tabular}{@{}c@{}}$\% \mathtt{Inf}\downarrow$ \end{tabular} & PSNR$\uparrow$ & SSIM$\uparrow$ & LPIPS$\downarrow$  \\ \midrule
LLFlow~\cite{wang2021low} & 20 & 20.51 & 0.897 & \multicolumn{1}{c|}{0.110} & 22 & 26.60 & \textbf{0.919} & 0.067\\
LLFlow\textsuperscript{\textdagger} & 20 & 19.66 & 0.894 & \multicolumn{1}{c|}{0.121} & 13 & 23.70 & 0.904 & 0.080\\
Ours & \textbf{0} & \textbf{21.00} & \textbf{0.904} & \multicolumn{1}{c|}{\textbf{0.106}} & \textbf{0} & \textbf{26.61} & \textbf{0.919} & \textbf{0.066}\\ \bottomrule
\end{tabular}
           \vspace{-.5em}
\caption{Quantitative comparison. 
The {\textdagger} sign denotes that the lower bound of the scale parameter is $0.1$.
`$\% \mathtt{Inf}$' refers to the percentage of conditional inputs that generate at least one $\mathtt{Inf}$ pixel out of 10 randomly generated latent codes, each with $\mathbf{z} \sim \mathcal{N}(\mathbf{0},\,\tau^2)$. 
The temperature of the latent code (\textit{i.e.}, $\tau$) is 1 for both LOL~\cite{Chen2018Retinex} and VE-LOL~\cite{ll_benchmark} datasets.}
   \vspace{-1em}
\label{table:llflow}
\end{table*}

\begin{figure}
    \centering
    \includegraphics[width=0.9\columnwidth]{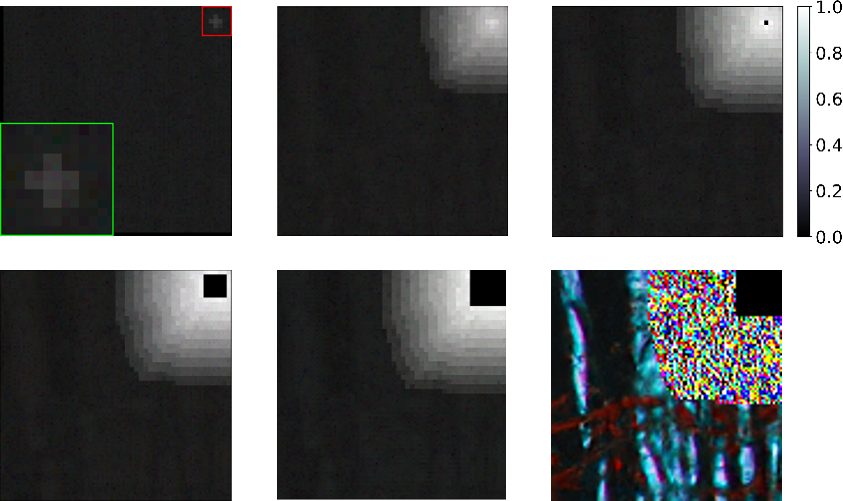}
               \vspace{-.5em}
    \caption{Visualization of feature map with exploding inverse (log scale). The  white pixels from the top right of the first image gradually spread out and eventually form a black region with $\mathtt{Inf}$ pixels.}
    \label{fig:feat}
      \vspace{-.5em}
\end{figure}

\section{Discussion}
\paragraph{Artifact type}
We could observe two types of artifacts. One shows random primary colors, while the other shows only black. To find out why those two types of artifacts appear, we extracted feature maps from the middle of the network (FS-NCSR~\cite{Song_2022_CVPR}) when they co-occurred. Figure \ref{fig:feat} shows the absolute values (log scale) of feature maps for a sample with both types of artifacts.
In the first feature map, which is the closest to the latent variable $\mathbf{z}$ among the five, it can be seen that the absolute value is large only in a very small area (zoomed). The bright pixels, which have large absolute values, gradually spread out for the rest of the feature maps and eventually form a black region with $\mathtt{Inf}$ values. Finally, both types of artifacts appear in the output, as in the last of Figure \ref{fig:feat}. 
This also explains why artifacts occur even without $\mathtt{Inf}$ pixels (see FS-NCSR\textsuperscript{\textdagger} in Figure \ref{tbl:table_of_figures} and Table \ref{table:fsncsr}).
One may wonder why the exploding inverse is gradually spreading, even though NFs do not use inter-pixel operations such as 3$\times$3 convolution or max-pooling. One reason is that NFs have the equivalent effect of using inter-pixel operations, employing both inter-channel and pixel shuffle operations. The other reason is that the $\mathrm{NN}$ used in \eqref{eq2.1.1}, \eqref{eq2.1.2} also employs inter-pixel operations.

\paragraph{Limitation}
Although our modified RQ-spline coupling transformation has an analytic inverse, it still imposes numerical overhead compared to the affine coupling transformation (about 2$\times$ training time).
As we mentioned in Section \ref{sec:rqspline}, there may exist a computationally efficient method to ensure robustness.
Measuring OOD scores is challenging but there is room for improvement for the accuracy of the Mahalanobis distance-based OOD score.

\section{Conclusion}
We addressed the issue of erroneous image samples in conditional NFs for inverse problems by revealing exploding inverse in affine coupling transformations and investigating OOD conditional inputs using the Mahalanobis distance. Then, we proposed the remarks to avoid exploding inverse in coupling transformations and suggested the modified RQ-spline coupling layer following the remarks for 2D toy, super-resolution space generation and low-light image enhancement, suppressing severe artifacts.

\section*{Acknowledgements}

This work was supported by the National Research Foundation of Korea(NRF) grants funded by the Korea government(MSIT) (NRF-2022R1A4A1030579) and Basic Science Research Program through the National Research Foundation of Korea(NRF) funded by the Ministry of Education(NRF-2017R1D1A1B05035810). Also, the authors acknowledged the financial supports from BK21 FOUR program of the Education and Research Program for Future ICT Pioneers, Seoul National University.

{\small
\bibliographystyle{ieee_fullname}
\bibliography{refs}

\begin{thebibliography}{10}\itemsep=-1pt

\bibitem{Abdelhamed_2019_ICCV}
Abdelrahman Abdelhamed, Marcus~A. Brubaker, and Michael~S. Brown.
\newblock Noise flow: Noise modeling with conditional normalizing flows.
\newblock In {\em ICCV}, 2019.

\bibitem{agustsson2017ntire}
Eirikur Agustsson and Radu Timofte.
\newblock Ntire 2017 challenge on single image super-resolution: Dataset and
  study.
\newblock In {\em CVPRW}, pages 126--135, 2017.

\bibitem{ardizzone2019guided}
Lynton Ardizzone, Carsten L{\"u}th, Jakob Kruse, Carsten Rother, and Ullrich
  K{\"o}the.
\newblock Guided image generation with conditional invertible neural networks.
\newblock {\em arXiv preprint arXiv:1907.02392}, 2019.

\bibitem{pmlr-v130-behrmann21a}
Jens Behrmann, Paul Vicol, Kuan-Chieh Wang, Roger Grosse, and Joern-Henrik
  Jacobsen.
\newblock Understanding and mitigating exploding inverses in invertible neural
  networks.
\newblock In {\em AISTATS}, pages 1792--1800, 2021.

\bibitem{DBLP:journals/corr/DinhKB14}
Laurent Dinh, David Krueger, and Yoshua Bengio.
\newblock {NICE:} non-linear independent components estimation.
\newblock In {\em {ICLR} (Workshop)}, 2015.

\bibitem{DBLP:conf/iclr/DinhSB17}
Laurent Dinh, Jascha Sohl{-}Dickstein, and Samy Bengio.
\newblock Density estimation using real {NVP}.
\newblock In {\em {ICLR} (Poster)}, 2017.

\bibitem{dolatabadi2020invertible}
Hadi~Mohaghegh Dolatabadi, Sarah Erfani, and Christopher Leckie.
\newblock Invertible generative modeling using linear rational splines.
\newblock In {\em AISTATS}, pages 4236--4246, 2020.

\bibitem{durkan2019neural}
Conor Durkan, Artur Bekasov, Iain Murray, and George Papamakarios.
\newblock Neural spline flows.
\newblock {\em NeurIPS}, 32, 2019.

\bibitem{durkan2019cubic}
Conor Durkan, Artur Bekasovs, Iain Murray, and Georgios Papamakarios.
\newblock Cubic-spline flows.
\newblock In {\em ICML Workshop on Invertible Neural Nets and Normalizing
  Flows}, 2019.

\bibitem{DBLP:conf/aaai/HeZ0LHY22}
Jinzheng He, Zhou Zhao, Yi Ren, Jinglin Liu, Baoxing Huai, and Nicholas~Jing
  Yuan.
\newblock Flow-based unconstrained lip to speech generation.
\newblock In {\em {AAAI}}, pages 843--851, 2022.

\bibitem{ho2019flow++}
Jonathan Ho, Xi Chen, Aravind Srinivas, Yan Duan, and Pieter Abbeel.
\newblock Flow++: Improving flow-based generative models with variational
  dequantization and architecture design.
\newblock In {\em ICML}, pages 2722--2730, 2019.

\bibitem{huang2015single}
Jia-Bin Huang, Abhishek Singh, and Narendra Ahuja.
\newblock Single image super-resolution from transformed self-exemplars.
\newblock In {\em CVPR}, pages 5197--5206, 2015.

\bibitem{jo2021srflow}
Younghyun Jo, Sejong Yang, and Seon~Joo Kim.
\newblock Srflow-da: Super-resolution using normalizing flow with deep
  convolutional block.
\newblock In {\em CVPR}, pages 364--372, 2021.

\bibitem{kim2021noise}
Younggeun Kim and Donghee Son.
\newblock Noise conditional flow model for learning the super-resolution space.
\newblock In {\em CVPRW}, pages 424--432, 2021.

\bibitem{kingma2014adam}
Diederik~P Kingma and Jimmy Ba.
\newblock Adam: A method for stochastic optimization.
\newblock In {\em ICLR}, 2015.

\bibitem{kingma2018glow}
Durk~P Kingma and Prafulla Dhariwal.
\newblock Glow: Generative flow with invertible 1x1 convolutions.
\newblock {\em NeurIPS}, 31, 2018.

\bibitem{kingma2016improved}
Durk~P Kingma, Tim Salimans, Rafal Jozefowicz, Xi Chen, Ilya Sutskever, and Max
  Welling.
\newblock Improved variational inference with inverse autoregressive flow.
\newblock {\em NIPS}, 29, 2016.

\bibitem{DBLP:journals/corr/KingmaW13}
Diederik~P. Kingma and Max Welling.
\newblock Auto-encoding variational bayes.
\newblock In {\em {ICLR}}, 2014.

\bibitem{kirichenko2020normalizing}
Polina Kirichenko, Pavel Izmailov, and Andrew~G Wilson.
\newblock Why normalizing flows fail to detect out-of-distribution data.
\newblock {\em NeurIPS}, 33, 2020.

\bibitem{kohler2021smooth}
Jonas K{\"o}hler, Andreas Kr{\"a}mer, and Frank No{\'e}.
\newblock Smooth normalizing flows.
\newblock {\em NeurIPS}, 34:2796--2809, 2021.

\bibitem{krizhevsky2009learning}
Alex Krizhevsky, Geoffrey Hinton, et~al.
\newblock Learning multiple layers of features from tiny images.
\newblock 2009.

\bibitem{lee2018simple}
Kimin Lee, Kibok Lee, Honglak Lee, and Jinwoo Shin.
\newblock A simple unified framework for detecting out-of-distribution samples
  and adversarial attacks.
\newblock {\em NeurIPS}, 31, 2018.

\bibitem{liang2021hierarchical}
Jingyun Liang, Andreas Lugmayr, Kai Zhang, Martin Danelljan, Luc Van~Gool, and
  Radu Timofte.
\newblock Hierarchical conditional flow: A unified framework for image
  super-resolution and image rescaling.
\newblock In {\em CVPR}, pages 4076--4085, 2021.

\bibitem{ll_benchmark}
Jiaying Liu, Xu Dejia, Wenhan Yang, Minhao Fan, and Haofeng Huang.
\newblock Benchmarking low-light image enhancement and beyond.
\newblock {\em IJCV}, 129:1153–1184, 2021.

\bibitem{liu2015deep}
Ziwei Liu, Ping Luo, Xiaogang Wang, and Xiaoou Tang.
\newblock Deep learning face attributes in the wild.
\newblock In {\em ICCV}, pages 3730--3738, 2015.

\bibitem{lu2020structured}
You Lu and Bert Huang.
\newblock Structured output learning with conditional generative flows.
\newblock In {\em AAAI}, pages 5005--5012, 2020.

\bibitem{Lucas2018review}
Alice Lucas, Michael Iliadis, Rafael Molina, and Aggelos~K. Katsaggelos.
\newblock Using deep neural networks for inverse problems in imaging: Beyond
  analytical methods.
\newblock {\em IEEE Signal Processing Magazine}, 35(1):20--36, 2018.

\bibitem{lugmayr2020srflow}
Andreas Lugmayr, Martin Danelljan, Luc~Van Gool, and Radu Timofte.
\newblock Srflow: Learning the super-resolution space with normalizing flow.
\newblock In {\em ECCV}, pages 715--732, 2020.

\bibitem{lugmayr2021ntire}
Andreas Lugmayr, Martin Danelljan, and Radu Timofte.
\newblock Ntire 2021 learning the super-resolution space challenge.
\newblock In {\em CVPRW}, pages 596--612, 2021.

\bibitem{lugmayr2022ntire}
Andreas Lugmayr, Martin Danelljan, Radu Timofte, Kang-wook Kim, Younggeun Kim,
  Jae-young Lee, Zechao Li, Jinshan Pan, Dongseok Shim, Ki-Ung Song, et~al.
\newblock Ntire 2022 challenge on learning the super-resolution space.
\newblock In {\em CVPRW}, pages 786--797, 2022.

\bibitem{lugmayr2022normalizing}
Andreas Lugmayr, Martin Danelljan, Fisher Yu, Luc Van~Gool, and Radu Timofte.
\newblock Normalizing flow as a flexible fidelity objective for photo-realistic
  super-resolution.
\newblock In {\em WACV}, pages 1756--1765, 2022.

\bibitem{muller2019neural}
Thomas M{\"u}ller, Brian McWilliams, Fabrice Rousselle, Markus Gross, and Jan
  Nov{\'a}k.
\newblock Neural importance sampling.
\newblock {\em ACM Transactions on Graphics (TOG)}, 38(5):1--19, 2019.

\bibitem{ongie2020review}
Gregory Ongie, Ajil Jalal, Christopher~A. Metzler, Richard~G. Baraniuk,
  Alexandros~G. Dimakis, and Rebecca Willett.
\newblock Deep learning techniques for inverse problems in imaging.
\newblock {\em IEEE Journal on Selected Areas in Information Theory},
  1(1):39--56, 2020.

\bibitem{papamakarios2021normalizing}
George Papamakarios, Eric~T Nalisnick, Danilo~Jimenez Rezende, Shakir Mohamed,
  and Balaji Lakshminarayanan.
\newblock Normalizing flows for probabilistic modeling and inference.
\newblock {\em Journal of Machine Learning Research}, 22(57):1--64, 2021.

\bibitem{papamakarios2017masked}
George Papamakarios, Theo Pavlakou, and Iain Murray.
\newblock Masked autoregressive flow for density estimation.
\newblock {\em NIPS}, 30, 2017.

\bibitem{prenger2019waveglow}
Ryan Prenger, Rafael Valle, and Bryan Catanzaro.
\newblock Waveglow: A flow-based generative network for speech synthesis.
\newblock In {\em IEEE ICASSP}, pages 3617--3621, 2019.

\bibitem{Pumarola_2020_CVPR}
Albert Pumarola, Stefan Popov, Francesc Moreno-Noguer, and Vittorio Ferrari.
\newblock C-flow: Conditional generative flow models for images and 3d point
  clouds.
\newblock In {\em CVPR}, 2020.

\bibitem{rezende2015variational}
Danilo Rezende and Shakir Mohamed.
\newblock Variational inference with normalizing flows.
\newblock In {\em ICML}, pages 1530--1538, 2015.

\bibitem{rezende2020normalizing}
Danilo~Jimenez Rezende, George Papamakarios, S{\'e}bastien Racaniere, Michael
  Albergo, Gurtej Kanwar, Phiala Shanahan, and Kyle Cranmer.
\newblock Normalizing flows on tori and spheres.
\newblock In {\em ICML}, pages 8083--8092, 2020.

\bibitem{Song_2022_CVPR}
Ki-Ung Song, Dongseok Shim, Kang-wook Kim, Jae-young Lee, and Younggeun Kim.
\newblock Fs-ncsr: Increasing diversity of the super-resolution space via
  frequency separation and noise-conditioned normalizing flow.
\newblock In {\em CVPRW}, pages 968--977, June 2022.

\bibitem{sukthanker2022generative}
Rhea~Sanjay Sukthanker, Zhiwu Huang, Suryansh Kumar, Radu Timofte, and Luc
  Van~Gool.
\newblock Generative flows with invertible attentions.
\newblock In {\em CVPR}, pages 11234--11243, 2022.

\bibitem{timofte2017ntire}
Radu Timofte, Eirikur Agustsson, Luc Van~Gool, Ming-Hsuan Yang, and Lei Zhang.
\newblock Ntire 2017 challenge on single image super-resolution: Methods and
  results.
\newblock In {\em CVPRW}, pages 114--125, 2017.

\bibitem{wang2022learning}
Haolin Wang, Jiawei Zhang, Ming Liu, Xiaohe Wu, and Wangmeng Zuo.
\newblock Learning diverse tone styles for image retouching.
\newblock {\em arXiv preprint arXiv:2207.05430}, 2022.

\bibitem{wang2018esrgan}
Xintao Wang, Ke Yu, Shixiang Wu, Jinjin Gu, Yihao Liu, Chao Dong, Yu Qiao, and
  Chen~Change Loy.
\newblock Esrgan: Enhanced super-resolution generative adversarial networks.
\newblock In {\em ECCVW}, September 2018.

\bibitem{wang2021low}
Yufei Wang, Renjie Wan, Wenhan Yang, Haoliang Li, Lap-Pui Chau, and Alex~C Kot.
\newblock Low-light image enhancement with normalizing flow.
\newblock In {\em AAAI}, pages 2604--2612, 2022.

\bibitem{Chen2018Retinex}
Chen Wei, Wenjing Wang, Wenhan Yang, and Jiaying Liu.
\newblock Deep retinex decomposition for low-light enhancement.
\newblock In {\em BMVC}, 2018.

\bibitem{wu2020stochastic}
Hao Wu, Jonas K{\"o}hler, and Frank No{\'e}.
\newblock Stochastic normalizing flows.
\newblock {\em NeurIPS}, 33:5933--5944, 2020.

\bibitem{wu2022bin}
Yiqiang Wu, Dapeng Tao, Yibing Zhan, and Chenyang Zhang.
\newblock Bin-flow: Bidirectional normalizing flow for robust image dehazing.
\newblock {\em IEEE Transactions on Image Processing}, 2022.

\bibitem{xiao2019learning}
Tong Xiao, Tian Xia, Yi Yang, Chang Huang, and Xiaogang Wang.
\newblock Learning from massive noisy labeled data for image classification.
\newblock In {\em CVPR}, pages 7320--7328, 2019.

\bibitem{zhang2015revisiting}
Kai Zhang, Xiaoyu Zhou, Hongzhi Zhang, and Wangmeng Zuo.
\newblock Revisiting single image super-resolution under internet environment:
  blur kernels and reconstruction algorithms.
\newblock In {\em Pacific Rim Conference on Multimedia}, pages 677--687.
  Springer, 2015.

\end{thebibliography}
}

\clearpage
\newpage
\appendix

\bigskip
\bigskip
\begin{center}
\textbf{\Large
 \quad \\ 
 Supplementary Material\\
On the Robustness of Normalizing Flows for Inverse Problems in Imaging}

\bigskip 
\bigskip 
{\large Seongmin Hong\textsuperscript{\rm 1}
\quad\quad
Inbum Park\textsuperscript{\rm 1}
\quad\quad\\
Se Young Chun\textsuperscript{\rm 1,2,*}\\
\textsuperscript{\rm 1}Dept. of Electrical and Computer Engineering,
\textsuperscript{\rm 2}INMC, Interdisciplinary Program in AI\\
Seoul National University, Republic of Korea\\
{\tt\small \{smhongok, inbum0215, sychun\}@snu.ac.kr}
}
\bigskip 
\bigskip 
 
\end{center}%
\renewcommand{\thesection}{S\arabic{section}}
\renewcommand{\thefigure}{S\arabic{figure}}
\renewcommand{\thetable}{S\arabic{table}}
\renewcommand{\theequation}{S\arabic{equation}}
\setcounter{equation}{0}
\setcounter{figure}{0}
\setcounter{table}{0}
\setcounter{page}{1}

\section{Experimental details and more results}
\subsection{OOD score}
\subsubsection{Super-resolution space generation}
We generated 1470 patches from the DIV2K~\cite{agustsson2017ntire} 4$\times$ validation dataset and ranked them based on their OOD score ($s_{\text{OOD}}$) using the conditioning network $g_{\boldsymbol{\theta}}$ of the fully-trained FS-NCSR~\cite{Song_2022_CVPR}.
For the super-resolution space generation, we concatenate the output of RRDB~\cite{wang2018esrgan} blocks 1, 8, 15, 22, instead of directly using the output of $g_{\boldsymbol{\theta}}$ for better feature representation of the conditioning encoder, which is trained on the DF2K~\cite{timofte2017ntire} training set 4$\times$. Then, we collect patches of size $160\times160$ and compute the OOD score (i.e. $s_{\text{OOD}}$) for each patch. 
The method of concatenating blocks of RRDB stems from the work of SRFlow~\cite{lugmayr2020srflow}, where they concatenate equally spaced RRDB blocks 1, 8, 15, 22, and 23 to obtain the final output of the conditioning encoder. This corresponds to Section 3.3 and Figure 4 of the main paper.

\paragraph{Pixel error}
To verify the presented OOD score, we computed the pixel error probability for each patch by generating 10 samples from each image of the DIV2K validation set. For each sample, we calculated the number of erroneous pixels, with the minimum and maximum error threshold set as $-0.5$ and $1.5$, respectively. This is because the output of the neural network should be within the range of $[0,1]$ before clamping. However, it is important to note that this pixel error is only a necessary condition for the exploding inverse, and not a necessary and sufficient condition. This is because a value of $0$ or $1$ obtained after clamping may be intended.
Table \ref{table:pixelerrors} shows the percentage of conditional inputs that generate at least one pixel whose value is outside the range of $[-0.5, 1.5]$. In the case of in-distribution, only 7\% of the conditional inputs generated at least one pixel error. However, in the case of OOD, 90\% of the conditional inputs generated pixel errors. It is worth noting that these values are significantly higher than the percentages of conditional inputs that generate an erroneous image in human eyes.
\begin{table}[]
\centering
\small
\begin{tabular}{cccc}
\toprule
            Train set           & Test set & Distribution & \% PixelErr$\downarrow$ \\
            \midrule
            
            \multirow{2}{*}{DF2K 4$\times$} & DIV2K 4$\times$ & \multicolumn{1}{c|}{in-distribution} & 7\%  \\
            & EUrban100 4$\times$ & \multicolumn{1}{c|}{OOD} & 90\% \\            
\bottomrule
\end{tabular}
           \vspace{-1em}
\caption{The percentage of conditional inputs that generate at least one error pixel (\textit{i.e.}, pixel value is out of $[-0.5, 1.5]$) out of 10 randomly generated latent codes, each with $\mathbf{z} \sim \mathcal{N}(\mathbf{0},\,\tau^2)$, where $\tau=0.9$.
}
\label{table:pixelerrors}
 \vspace{-1em}
\end{table}

\paragraph{Enhanced Urban100}
To investigate the case of severe OOD, we made modifications to the Urban100 dataset~\cite{huang2015single}. The original Urban100 dataset has an average OOD score of $\mathbb{E}[s_{\text{OOD}}] = 331.01$, which is only slightly larger than that of the training set ($\mathbb{E}[s_{\text{OOD}}] = 236.78$). To generate a severe OOD dataset, we enhanced each image of the Urban100 dataset by strengthening the high frequency components using a convolution kernel $\mathbf{H}$, where
\begin{equation}
    \mathbf{H} = \frac{1}{3}\left[\begin{matrix}
        -1 & -4 & -1 \\
        -4 & 26 & -4 \\
        -1 & -4 & -1 \\
    \end{matrix}\right]. 
\end{equation}
This operation enhanced the OOD score to $\mathbb{E}[s_{\text{OOD}}] = 511.35$, which is much larger than that of the training set.

\subsubsection{Low-light image enhancement}
\begin{figure}
    \centering
    \includegraphics[width=0.5\textwidth]{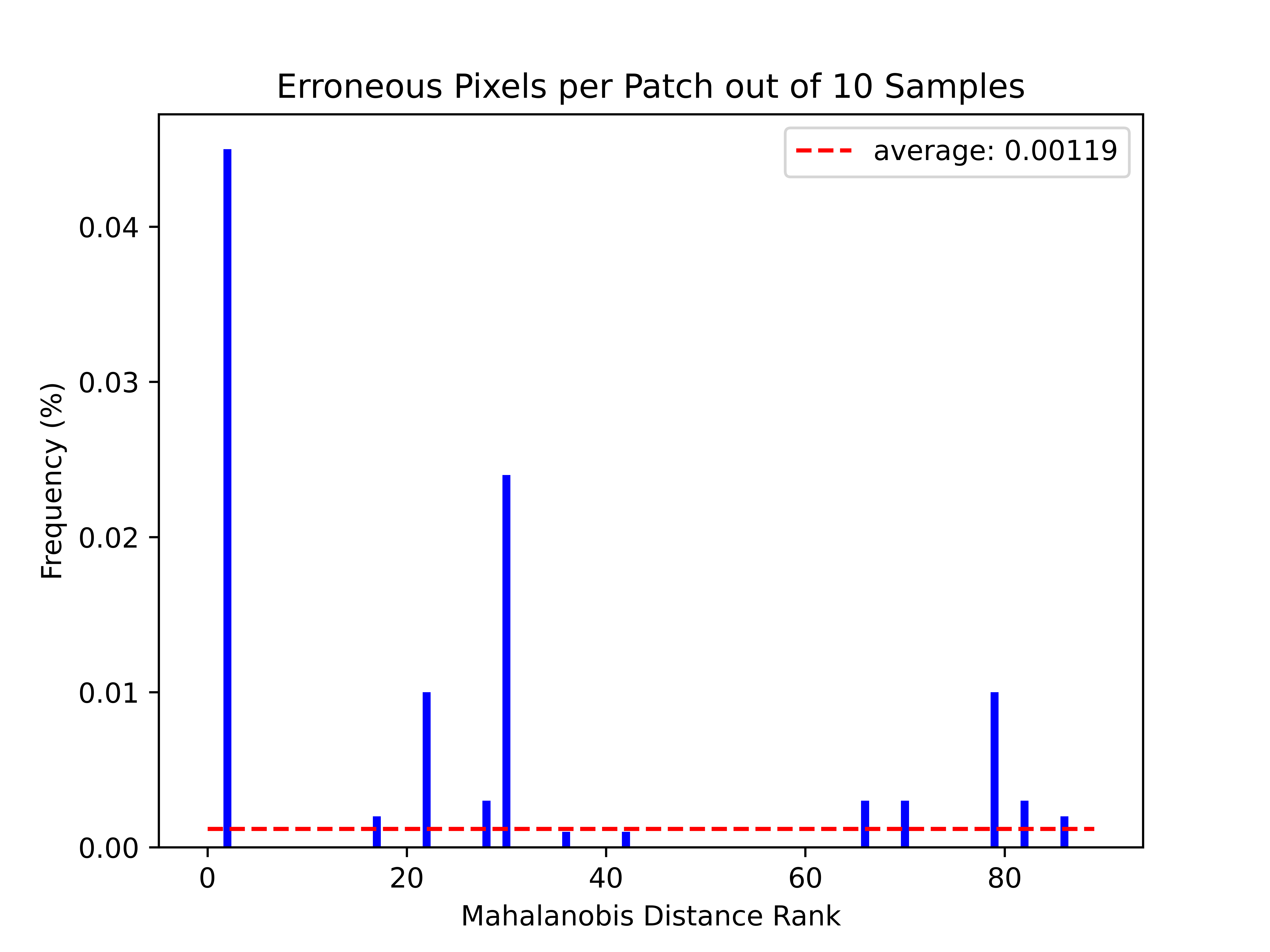}
    \caption{Pixel error probability for the patches ranked according to their OOD score ($s_{\text{OOD}}$). The average of 90 patches is marked as a dashed horizontal line.}
    \label{fig:llflow_ood}
\end{figure}

We also calculate the OOD score for the low-light image enhancement on the LOL~\cite{Chen2018Retinex} testset.  The second row of Figure 1 in the main paper shows an erroneous sample generated from the patch with the highest OOD score among the 90 patches.
Similar to the task of super-resolution space generation, we concatenate the output of RRDB blocks 1, 3, 5, 7 as the output of $g_{\boldsymbol{\theta}}$, the conditioning network fully trained on the LOL~\cite{Chen2018Retinex} training set. Then, we collect a total of 90 patches, each of size $100\times100$. We rank the OOD score based on the mahalanobis score of each patch. In Figure \ref{fig:llflow_ood}, we show the pixel error probability of the LOL dataset ranked according to the OOD score of each patch.
In all cases, ours showed the best results.

\subsection{2D toy experiment}\label{sec:s1.2}
\paragraph{Training data} The training data is obtained by the following equation:
\begin{equation}\label{eq1.2.1}
    \mathbf{x}=
    \left[
    \begin{matrix}
        x_1 \\
        x_2 \\
    \end{matrix}
    \right]
    =
    \left[
    \begin{matrix}
        \frac{\sqrt{3}}{4} & -\frac{1}{10} \\
        \frac{1}{4} & -\frac{\sqrt{3}}{10} \\
    \end{matrix}
    \right]
    \left[
    \begin{matrix}
        u_1 \\
        u_2 \\
    \end{matrix}
    \right],
\end{equation}
where $u_1, u_2 \sim \mathcal{U}(-1,1)$ (\textit{i.e.}, uniform distribution on $(-1,1)$).
We generated 100,000 samples using \eqref{eq1.2.1}.

\paragraph{Network architecture}
$\mathrm{NN}$ in the coupling layers was a fully connected network composed of four hidden layers with a width of 64. For the modified RQ-spline coupling layer, the output of $\mathrm{NN}$ is four-dimensional (\textit{i.e.}, $\mathbf{h}_2 \in \mathbb{R}^4$). The four components of the output are bias (\textit{i.e.} $t$), input coordinate of the learnable knot, output coordinate of the learnable knot, and slope of the learnable knot (\textit{i.e.}, derivative of the RQ-spline transformation at the learnable knot). The input coordinate of the learnable knot is normalized (via sigmoid) to be in $(B_1 + \epsilon, B_2 - \epsilon)$, and the output coordinate of the learnable knot is normalized (via sigmoid) to be in $(B_1 + t + \epsilon, B_2 + t - \epsilon)$. We set the slope of the learnable knot in $(\epsilon, \infty)$, via exponential function. We used $(B_1, B_2) = (-0.5, 0.5)$ and $\epsilon = 0.001$. $\mathbf{z}$ was assumed to be the standard Gaussian.

\paragraph{Training}
We trained the network using Adam optimizer~\cite{kingma2014adam}, with $(\beta_1,\beta_2) = (0.9, 0.999)$, learning rate $5 \times 10^{-4}$, batch size 1,000, for 8,000 iterations.

\begin{figure}
    \centering
        \includegraphics[width=0.5\textwidth]{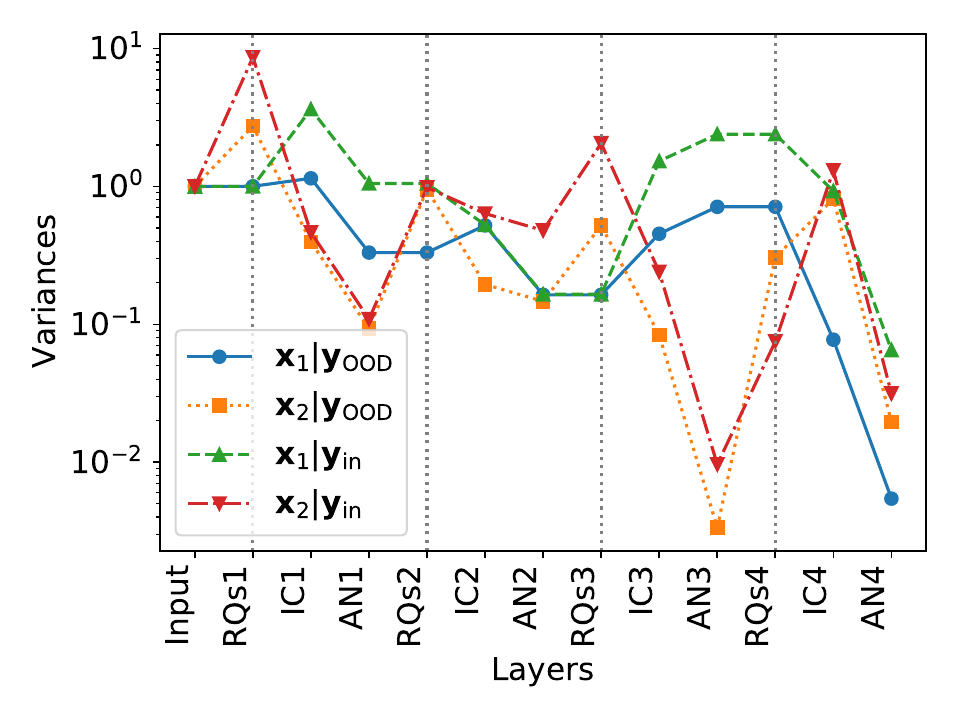} 
    \caption{Variances of features for in-distribution and OOD conditional inputs. \textsf{RQs}, \textsf{IC} and \textsf{AN} denote the conditional RQ-spline coupling, invertible 1$\times$1 convolution and activation normalization layers, respectively.}
    \label{fig:s2dToyResults}
\end{figure}

\paragraph{Additional results}
Figure \ref{fig:s2dToyResults} shows the variances of features in each layer of the flow model, employing RQ-spline coupling layer, where the conditional inputs are in-distribution or OOD. Unlike the results of 
the affine coupling layer in the main text, 
variances for OOD conditional inputs do not explode.

\subsection{Super-resolution space generation}

\paragraph{Training data}
For the DIV2K~\cite{agustsson2017ntire} validation set, the training set is a combination of DIV2K 1-800 and Flickr2K~\cite{timofte2017ntire} 1-2,650 (total 3,450, and the union of DIV2K and Flickr2K is referred as DF2K), and the test set is DIV2K 801-900 and EUrban100. We used 160$\times$160 RGB patches as HR images. We randomly cropped the original images to generate 160$\times$160 RGB patches. We used bicubic kernel to generate the conditional inputs. We applied 90$^{\circ}$ rotations and horizontal flips randomly for data augmentation.

\paragraph{Network architecture} In the case of substituting the affine coupling layer of FS-NCSR~\cite{Song_2022_CVPR} with the modified RQ-spline coupling layer, the output of $\mathrm{NN}$ was set in the same manner as in Section \ref{sec:s1.2}. $\mathrm{NN}$ was a CNN, which is the same as FS-NCSR. The other structures were also exactly same as FS-NCSR. We set $\tau = 0.9$.

\paragraph{Training}
We trained the network using Adam optimizer~\cite{kingma2014adam}, with $(\beta_1,\beta_2) = (0.9, 0.999)$, initial learning rate $2\times 10^{-4}$. The learning rate is halved when 50\%, 75\%, 90\%, 95\% of the total number of iterations are trained. For DIV2K 4$\times$  dataset, batch size was 16, and the number of iterations was 180,000. For DIV2K 8$\times$ dataset, batch size was 12, and the number of iterations was 200,000. 
For fast training on 8$\times$ datasets, we replaced the invertible 1$\times$1 convolutions with fixed random unitary matrices. This technique is proposed by Lugmayr \textit{et al.}~\cite{lugmayr2022normalizing}, and has the effect of reducing training time while maintaining performance. We train the networks on a NVIDIA GeForce GTX 3090 GPU.

\begin{figure*}
    \centering
    \begin{tabular}{ccccc}
        \includegraphics[width=0.175\textwidth]{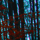} &
        \includegraphics[width=0.175\textwidth]{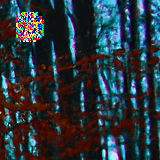} &
        \includegraphics[width=0.175\textwidth]{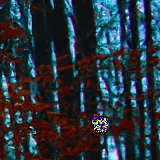} &
        \includegraphics[width=0.175\textwidth]{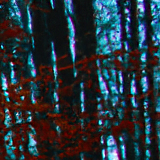} &
        \includegraphics[width=0.175\textwidth]{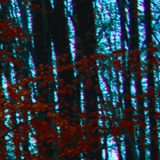} \\
        \includegraphics[width=0.175\textwidth]{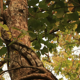} &
        \includegraphics[width=0.175\textwidth]{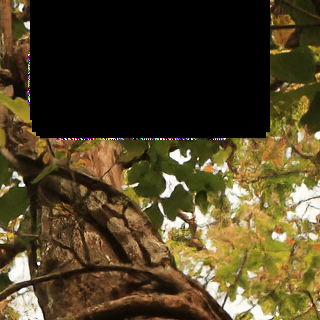} &
        \includegraphics[width=0.175\textwidth]{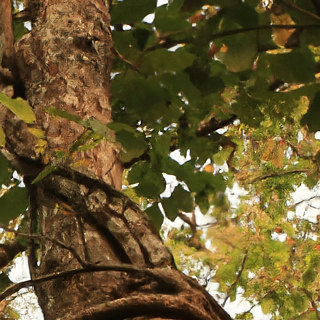} &
        \includegraphics[width=0.175\textwidth]{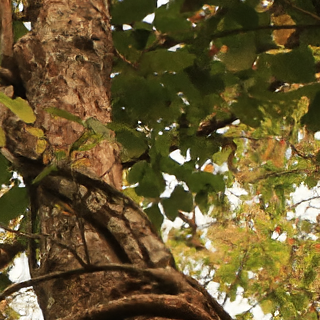} &
        \includegraphics[width=0.175\textwidth]{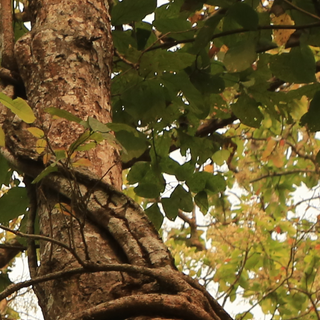} \\
        \includegraphics[width=0.175\textwidth]{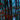} &
        \includegraphics[width=0.175\textwidth]{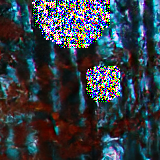} &
        \includegraphics[width=0.175\textwidth]{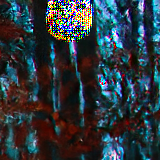} &
        \includegraphics[width=0.175\textwidth]{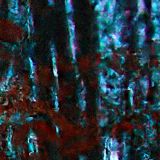} &
        \includegraphics[width=0.175\textwidth]{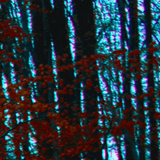} \\
        \includegraphics[width=0.175\textwidth]{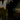} &
        \includegraphics[width=0.175\textwidth]{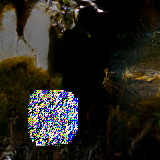} &
        \includegraphics[width=0.175\textwidth]{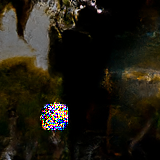} &
        \includegraphics[width=0.175\textwidth]{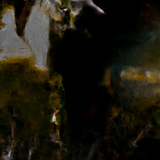} &
        \includegraphics[width=0.175\textwidth]{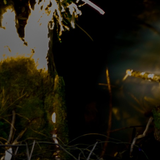} \\
        \includegraphics[width=0.175\textwidth]{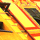} &
        \includegraphics[width=0.175\textwidth]{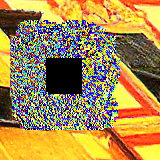} &
        \includegraphics[width=0.175\textwidth]{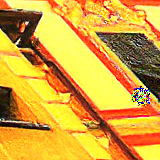} &
        \includegraphics[width=0.175\textwidth]{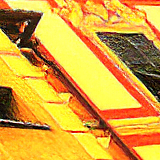} &
        \includegraphics[width=0.175\textwidth]{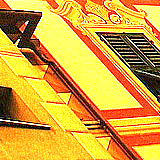} \\
        \includegraphics[width=0.175\textwidth]{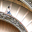} &
        \includegraphics[width=0.175\textwidth]{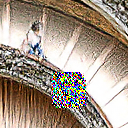} &
        \includegraphics[width=0.175\textwidth]{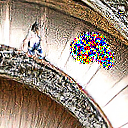} &
        \includegraphics[width=0.175\textwidth]{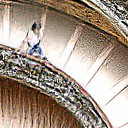} &
        \includegraphics[width=0.175\textwidth]{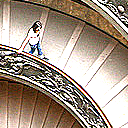} \\
        
        LR & FS-NCSR~\cite{Song_2022_CVPR} & FS-NCSR\textsuperscript{\textdagger} & Ours & Ground Truth\\
    \end{tabular}
               \vspace{-1em}
    \caption{Qualitative comparison of coupling transformation in super-resolution space generation. The 1st-2nd, 3rd-4th, and 5th-6th rows show the samples from DIV2K~\cite{agustsson2017ntire} 4$\times$, DIV2K 8$\times$, and EUrban100 4$\times$. The {\textdagger} sign denotes that the lower bound of the scale parameter is $0.1$.}
    \label{fig:fsncsr_supp}
\end{figure*}

\paragraph{Additional results} We provide additional examples of artifacts in Figure \ref{fig:fsncsr_supp}.

\subsubsection{Additional experiment on another dataset}
\begin{figure*}
    \centering
    \begin{tabular}{ccccc}
        \includegraphics[width=0.175\textwidth]{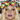} &
        \includegraphics[width=0.175\textwidth]{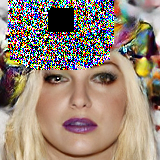} &
        \includegraphics[width=0.175\textwidth]{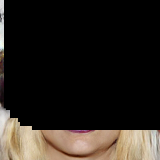} &
        \includegraphics[width=0.175\textwidth]{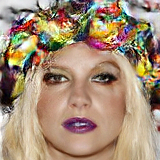} &
        \includegraphics[width=0.175\textwidth]{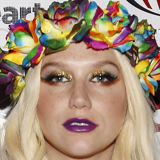} \\   
        \includegraphics[width=0.175\textwidth]{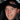} &
        \includegraphics[width=0.175\textwidth]{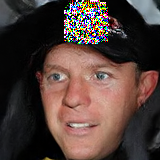} &
        \includegraphics[width=0.175\textwidth]{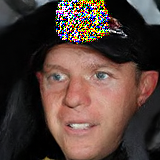} &
        \includegraphics[width=0.175\textwidth]{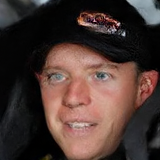} &
        \includegraphics[width=0.175\textwidth]{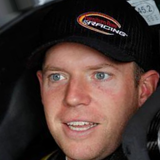} \\
        \includegraphics[width=0.175\textwidth]{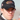} &
        \includegraphics[width=0.175\textwidth]{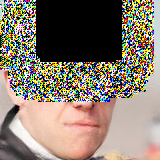} &
        \includegraphics[width=0.175\textwidth]{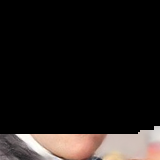} &
        \includegraphics[width=0.175\textwidth]{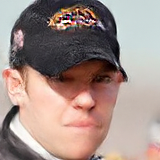} &
        \includegraphics[width=0.175\textwidth]{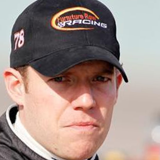} \\    
        LR & FS-NCSR~\cite{Song_2022_CVPR} & FS-NCSR\textsuperscript{\textdagger} & Ours & Ground Truth\\
    \end{tabular}
               \vspace{-1em}
    \caption{Qualitative comparison of coupling transformation in super-resolution space generation, on CelebA~\cite{liu2015deep} 8$\times$. The {\textdagger} sign denotes that the lower bound of the scale parameter is $0.1$.}
    \label{fig:fsncsr_supp_celeba}
\end{figure*}

For the CelebA~\cite{liu2015deep} validation set, CelebA 1-182,340 served as the training set, while CelebA 182,341-202,600 (total 20,260) was the validation set.
In the case of CelebA 8$\times$ dataset, batch size was 12, and the number of iterations was 100,000. We provide examples of artifacts in Figure \ref{fig:fsncsr_supp_celeba}.
Table \ref{table:celebA} shows the quantitative results of 8x super-resolution space generation on CelebA datasets. $\% \mathtt{Inf}$ demonstrates that our method effectively suppressed exploding inverses. However, compared to Table 1 in the text, FS-NCSR has a relatively small $\% \mathtt{Inf}$, as the CelebA dataset has very few OOD conditional inputs overall. Since the occurrence of exploding inverses was infrequent in this dataset, there was no significant difference in $\overline{\min}$ and $\overline{\sigma}$. Nevertheless, our method, with the exception of $\overline{\min}$, exhibited the most favorable results.

\begin{table}[]
\centering
\small
\begin{tabular}{lcccc}
\toprule
             Train $\rightarrow$ Test              & \multicolumn{4}{c}{CelebA 8$\times$ $\rightarrow$ CelebA 8$\times$} \\
           
\multicolumn{1}{l|}{Model} &  \multicolumn{1}{c}{$\% \mathtt{Inf}\downarrow$} & $\overline{\min}\uparrow$ & \multicolumn{1}{c|}{$\overline{\sigma}\downarrow$} & $\%$ PixelErr $\downarrow$\\ \midrule
FS-NCSR~\cite{Song_2022_CVPR}                       &   0.074 & 50.73 & 0.223 & 2.78 \\
FS-NCSR\textsuperscript{\textdagger}                    & 0.020 & \textbf{51.09} & 0.214 & 1.62  \\
Ours                       &  \textbf{0} & 50.63 & \textbf{0.199} & \textbf{0.48} \\ \bottomrule
\end{tabular}
           \vspace{-.5em}
\caption{Quantitative comparison on CelebA 8$\times$ dataset. The {\textdagger} sign denotes that the lower bound of the scale parameter is $0.1$.
`$\% \mathtt{Inf}$' and `$\%$ PixelErr' refer to the percentage of conditional inputs that generate at least one $\mathtt{Inf}$ pixel / pixel whose value is out of $[-0.5, 1.5]$ out of 10 randomly generated latent codes, each with $\mathbf{z} \sim \mathcal{N}(\mathbf{0},\,\tau^2)$, respectively. $\overline{\min}$ and $\overline{\sigma}$ refer the average of the minimum and standard deviation of LR-PSNR, respectively.}
\label{table:celebA}
 \vspace{-1em}
\end{table}

\subsection{Low-light enhancement}

\paragraph{Sampling method} LLFlow~\cite{wang2021low} suggested two sampling schemes to solve the low-light image enhancement problem. One is to fix the latent code $\mathbf{z}$ to $\mathbf{0}$ (\textit{i.e.}, $\hat{\mathbf{x}} = f_{\boldsymbol{\theta}}^{-1}(\mathbf{0};\mathbf{y})$). The other is to select a batch of $\mathbf{z}$ from the Gaussian distribution, and then calculate the mean (\textit{i.e.}, $\hat{\mathbf{x}} = \mathbb{E}_{\mathbf{z} \sim \mathcal{N}(0,\tau^2)}[ f_{\boldsymbol{\theta}}^{-1}(\mathbf{z};\mathbf{y})]$). Although LLFlow proposed both schemes, the authors only experimented with the first scheme. Here, we show experimental results that the second scheme generates erroneous images, while our solution does not.

\paragraph{Training data} We follow the training method of LLFlow~\cite{wang2021low} where we perform two evaluations: one on the LOL~\cite{Chen2018Retinex} validation set (trained on the LOL training set) and one on the VE-LOL~\cite{ll_benchmark} captured validation set (trained on the LOL training set).

\paragraph{Training} We trained the network using the same hyperparameters as the authors of LLFlow~\cite{wang2021low}. For both experiments, the batch size was 16 for the baseline model and 8 for our model. The number of iterations was 40,000 for the baseline model and 80,000 for our model. We train the networks on a NVIDIA Titan RTX GPU.

\begin{figure*}
    \centering
    \begin{tabular}{ccccc}
        \includegraphics[width=0.175\textwidth]{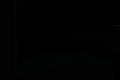} & 
        \includegraphics[width=0.175\textwidth]{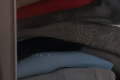} &
        \includegraphics[width=0.175\textwidth]{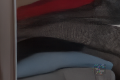} &
        \includegraphics[width=0.175\textwidth]{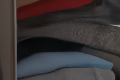} &
        \includegraphics[width=0.175\textwidth]{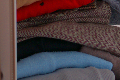} \\
        \includegraphics[width=0.175\textwidth]{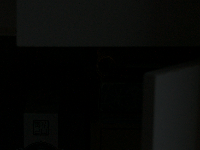} & 
        \includegraphics[width=0.175\textwidth]{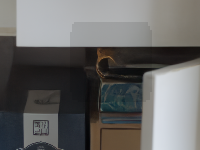} &
        \includegraphics[width=0.175\textwidth]{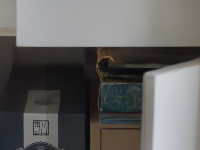} &
        \includegraphics[width=0.175\textwidth]{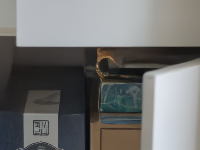} &
        \includegraphics[width=0.175\textwidth]{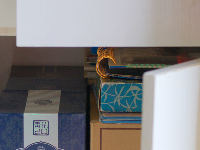} \\
        \includegraphics[width=0.175\textwidth]{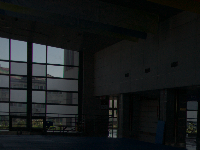} & 
        \includegraphics[width=0.175\textwidth]{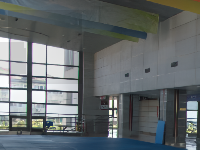} &
        \includegraphics[width=0.175\textwidth]{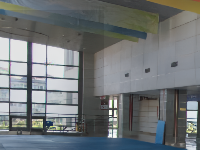} &
        \includegraphics[width=0.175\textwidth]{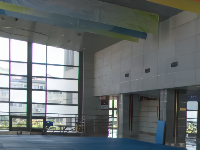} &
        \includegraphics[width=0.175\textwidth]{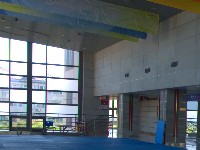} \\
        Input & 
        LLFlow~\cite{wang2021low} & LLFlow\textsuperscript{\textdagger} & Ours & Ground Truth\\
    \end{tabular}
    \caption{Qualitative comparison of coupling transformation in low-light image enhancement on the LOL~\cite{Chen2018Retinex} dataset. The {\textdagger} sign denotes that the lower bound of the scale parameter is 0.1.}
    \label{fig:llflow_supp_lol}
\end{figure*}

\begin{figure*}
    \centering
    \begin{tabular}{ccccc}
        \includegraphics[width=0.175\textwidth]{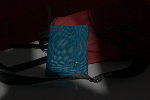} & 
        \includegraphics[width=0.175\textwidth]{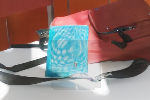} &
        \includegraphics[width=0.175\textwidth]{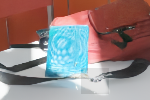} &
        \includegraphics[width=0.175\textwidth]{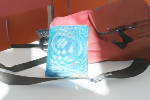} &
        \includegraphics[width=0.175\textwidth]{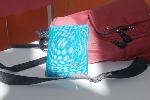} \\
        \includegraphics[width=0.175\textwidth]{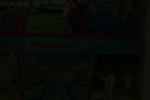} & 
        \includegraphics[width=0.175\textwidth]{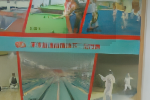} &
        \includegraphics[width=0.175\textwidth]{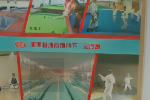} &
        \includegraphics[width=0.175\textwidth]{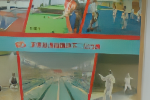} &
        \includegraphics[width=0.175\textwidth]{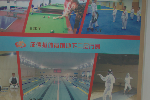} \\
        \includegraphics[width=0.175\textwidth]{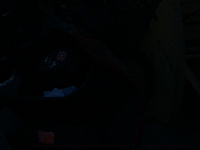} & 
        \includegraphics[width=0.175\textwidth]{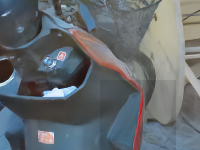} &
        \includegraphics[width=0.175\textwidth]{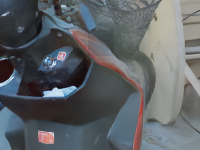} &
        \includegraphics[width=0.175\textwidth]{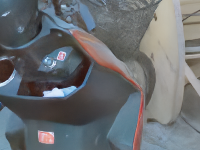} &
        \includegraphics[width=0.175\textwidth]{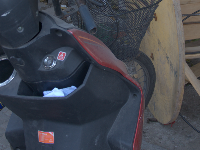} \\
        Input & 
        LLFlow~\cite{wang2021low} & LLFlow\textsuperscript{\textdagger} & Ours & Ground Truth\\
    \end{tabular}
    \caption{Qualitative comparison of coupling transformation in low-light image enhancement on the VE-LOL~\cite{ll_benchmark} dataset. The {\textdagger} sign denotes that the lower bound of the scale parameter is 0.1.}
    \label{fig:llflow_supp_ve}
\end{figure*}

\paragraph{Additional results} We provide additional examples of artifacts in Figures \ref{fig:llflow_supp_lol} and \ref{fig:llflow_supp_ve}.

\section{Additional Resources}
We used the source code of Zhang \textit{et al.}~\cite{zhang2015revisiting} to zoom images.

\end{document}